\documentclass[10pt,twocolumn,letterpaper]{article}

\usepackage[pagenumbers]{cvpr} 

\definecolor{cvprblue}{rgb}{0.21,0.49,0.74}
\usepackage[pagebackref,breaklinks,colorlinks,allcolors=cvprblue]{hyperref}
\usepackage{amsfonts}       
\usepackage{nicefrac}       
\usepackage{microtype}      
\usepackage{xcolor}         
\usepackage{amsmath}
\usepackage{graphicx}
\usepackage{multirow}
\usepackage{soul}
\usepackage{adjustbox}
\usepackage{tikz}
\usepackage{circuitikz}
\usepackage[table]{xcolor}
\usepackage{subcaption}
\definecolor{rankone}{RGB}{255,235,153}    
\definecolor{ranktwo}{RGB}{224,224,224}    
\definecolor{rankthree}{RGB}{230,204,178}  
\DeclareMathOperator*{\argmax}{argmax}

\definecolor{mynicegreen}{RGB}{98,179,104}
\usepackage{pifont}           

\definecolor{grpoblue}{RGB}{31,79,191}\definecolor{grtogreen}{RGB}{26,138,58}   
\newsavebox{\tabboxA}\newsavebox{\tabboxB}\newsavebox{\tabboxC}   
\def\confName{CVPR}
\def\confYear{2027}

\title{GRTO: Group Relative Tool Optimization for Referring Segmentation}

\author{%
    Mario Markov{$^{\dagger}$},\quad
    Stefan Maria Ailuro,\quad
    Mohammad Mahdi,\\
    Luc Van Gool,\quad
    Danda Pani Paudel \\
    INSAIT, Sofia University ``St. Kliment Ohridski''\\
    {$^{\dagger}$ \tt\small mario.markov@insait.ai} \\
}

\begin{document}

\twocolumn[{%
    \begin{@twocolumnfalse}
        \maketitle
\vspace{-5mm}
        \begin{minipage}{\textwidth}
\def\teaserscale{1.0}

    \captionsetup{type=figure}
    \begin{minipage}[t]{0.715\dimexpr\teaserscale\textwidth\relax}
        \vspace{0pt}
        \begin{subfigure}{\linewidth}
            \centering
            \includegraphics[width=\linewidth]{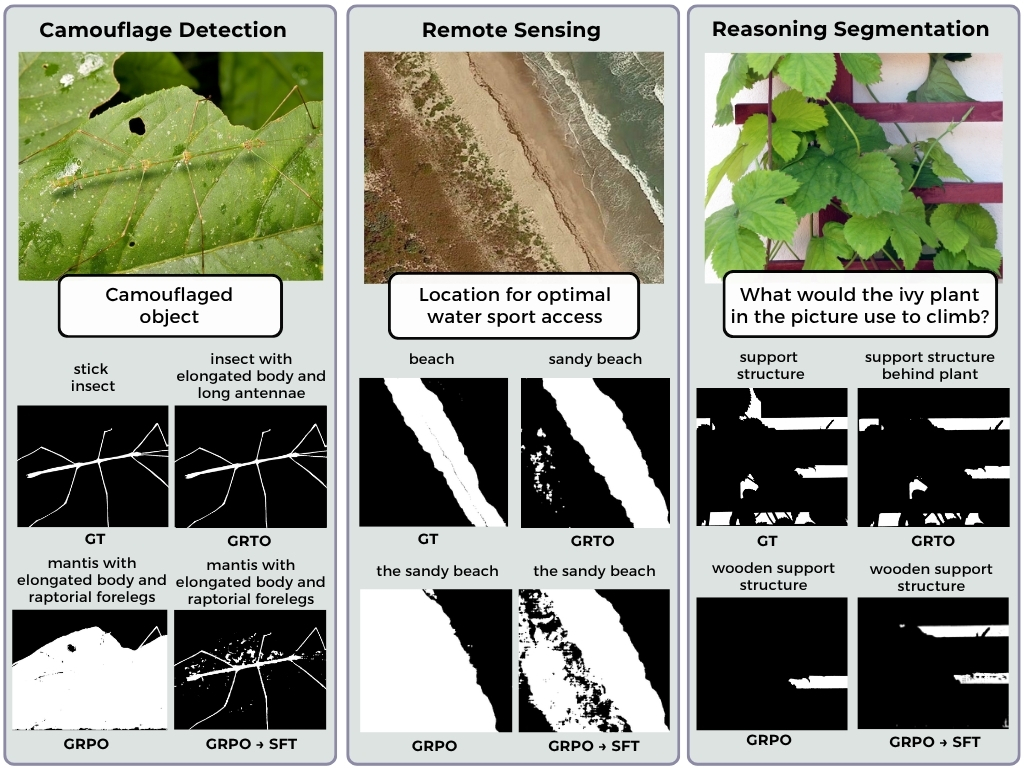}
            \caption{Referring Segmentation examples and results in three benchmarks.}
            \label{fig:ta}
        \end{subfigure}
    \end{minipage}\hspace{0.01875\dimexpr\teaserscale\textwidth\relax}
    \begin{minipage}[t]{0.27\dimexpr\teaserscale\textwidth\relax}
        \vspace{0pt}
        \captionsetup[subfigure]{justification=centering}
        \begin{subfigure}{\linewidth}
            \centering
            \includegraphics[width=\linewidth]{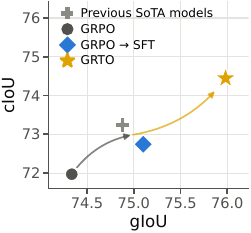}
            \caption{GRTO outperforms SoTA methods.}
            \label{fig:tb}
        \end{subfigure}

       \vspace{1mm}
        \begin{subfigure}{\linewidth}
            \centering
            \includegraphics[width=\linewidth]{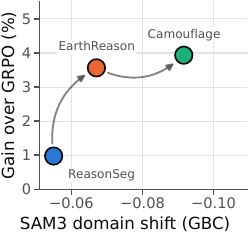}
            \caption{GRTO's gains with tool domain shift.}
            \label{fig:tc}
        \end{subfigure}
    \end{minipage}
\vspace{-2mm}
\setcounter{figure}{0}
    \captionof{figure}{(a) Examples from three benchmarks requiring simultaneous reasoning and segmentation. Here, GRPO uses a frozen tool, whereas GRPO$\rightarrow$SFT finetunes the tool naively afterwards. The proposed joint optimization of GRTO yields the best results. b) gIoU and cIoU averaged over the three domains. GRTO is superior to GRPO, GRPO$\rightarrow$SFT, and the previous state-of-the-art methods. c) GRTO's relative gain over GRPO grows with the domain shift of the frozen SAM3, measured by the Gaussian Bhattacharyya coefficient (GBC)~\cite{pandy2022transferability} between target and background pixels in frozen SAM3  features. All results are averaged over each domain's test sets and two seeds.}
    \label{fig:teaser}
        \end{minipage}
        \vspace{1em}
    \end{@twocolumnfalse}
}]

\maketitle

\begin{abstract}
Referring segmentation requires reasoning to connect language with visual evidence and identify the intended target in challenging scenarios. This makes GRPO a natural choice for training VLMs together with a segmentation tool. However, existing approaches optimize only the VLM, assuming that the tool is perfect and leaving it untouched. This assumption is particularly limiting when the tool faces domain shift, as segmentation errors can undermine even correct reasoning. We therefore argue that the segmentation tool should be optimized together with the reasoning policy. However, this is challenging because policy optimization and differentiable tool learning rely on fundamentally different objectives.
In this work, we introduce group relative tool optimization (GRTO), a mathematically grounded two-stage framework for jointly optimizing a policy and its differentiable tool. The key idea is to first align the tool with the policy expected after joint training, rather than the initial policy. The prealigned tool then leads to a more robust reward signal and superior policy convergence in subsequent policy training. In the first stage, GRTO bootstraps the tool toward the theoretically optimal policy, capturing readily attainable gains without costly on-policy rollouts. This alignment bootstraps the second stage, where GRTO extends GRPO to jointly refine the policy and tool. Across three challenging referring segmentation benchmarks, GRTO achieves substantial improvements over plain GRPO and alternative tool training procedures, matching or surpassing domain-specific state-of-the-art methods. This demonstrates the value of unifying reinforcement learning with differentiable auxiliary objectives for reasoning-intensive segmentation.
\end{abstract}

\section{Introduction}
\label{sec:intro}

Referring segmentation requires a model to produce a pixel-level mask for a target specified by natural language. Unlike conventional semantic segmentation, it conditions the output on an open-ended prompt. Flexible target descriptions are central to open-ended visual interfaces, robotics, medical image analysis, and safety-critical perception, where segmentation should follow intent rather than fixed categories~\cite{jiang2024attnibvs,ding2025multimodalsurvey,10.1145/3746027.3755166}. In many of these scenarios, the model must reason about the intended target using contextual and semantic visual evidence before producing its mask. Thus, referring segmentation requires two complementary capabilities: reasoning to identify the intended target and accurate segmentation to delineate it, as shown in Figure~\ref{fig:teaser}(a). Learning the former is particularly challenging because the intended target interpretation is not explicitly specified.

This makes GRPO a natural choice for training VLMs together with a segmentation tool~\cite{liu2025seg,liu2025visionreasoner,yun2026star}. By using final-mask rewards, GRPO allows VLMs to explore alternative interpretations and reinforce those leading to better segmentation results. However, existing approaches optimize only the VLM while keeping the segmentation tool frozen, implicitly assuming that the tool is capable enough. This assumption is problematic under domain shift, where tool errors can undermine even correct reasoning by producing poor masks and misleading rewards. A naive alternative is to first train the reasoning policy with GRPO using the frozen tool and then fine-tune the tool afterwards. This sequential strategy is suboptimal, as shown in Figure~\ref{fig:teaser}(b). Ideally, policy and tool should be optimized jointly so that improvements in either component benefit the other. In fact, such joint learning has shown benefits across various domains~\cite{guu2020realm,lewis2020rag,izacard2023atlas,bahar2021tight,moritz2025transcribing,chen2024scaling}.
Joint optimization, however, is challenging because policy optimization relies on costly on-policy sampling and reward-based objectives, whereas tool learning uses differentiable segmentation objectives. Reconciling these fundamentally different optimization signals is therefore key to effective joint policy-tool optimization.

To address these challenges, we introduce group relative tool optimization (GRTO), a mathematically grounded two-stage framework for jointly optimizing a reasoning policy and its differentiable segmentation tool. The key insight is to align the tool with the policy expected after joint optimization, rather than the initial policy. In the first stage, bootstrapped tool optimization (BTO) trains the tool toward the analytically optimal policy, capturing readily attainable improvements without the costly on-policy rollouts. This prealignment provides a stronger tool and more informative reward signals for the second stage, joint policy-tool optimization (JPTO), which extends GRPO~\cite{shao2024deepseekmath} to jointly refine the reasoning policy and the tool. By bootstrapping tool learning before joint optimization, GRTO reconciles the fundamentally different objectives of reinforcement learning and differentiable tool learning, leading to more effective policy optimization and a more robust final system.


We evaluate GRTO across three challenging benchmarks: camouflaged object detection, remote sensing segmentation, and reasoning segmentation. As shown in Figure~\ref{fig:teaser}, GRTO consistently improves over both policy-only GRPO and sequential tool fine-tuning (GRPO$\rightarrow$SFT), achieving the best overall performance and matching or surpassing domain-specific best methods. Notably, its gains over GRPO increase with the domain shift of the frozen segmentation tool (Figure~\ref{fig:teaser}(c)), demonstrating the importance of tool adaptation. These results highlight the benefits of jointly optimizing reasoning and segmentation, particularly when the initial tool is not sufficiently capable across domains.

\section{Related works}

\begin{figure}[t]
    \centering
    \includegraphics[width=\linewidth]{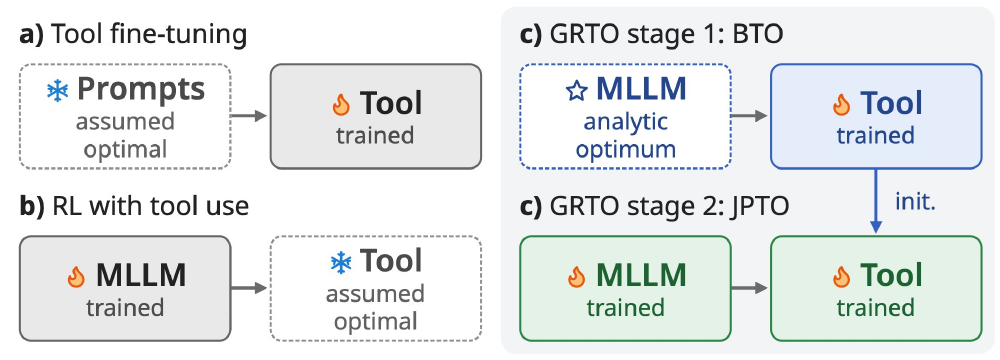}
    \caption{a) Most tool fine-tuning methods require instruction prompts to be perfectly specified and/or adjusted for the tool~\cite{carion2025sam3, lai2023lisa, hu2024rsvp, mei2023camouflaged, zhao2024focusdiffuser, sun2024frequency, he2023camouflaged, jan2025c3net, fan2021cod10k}. b) Referring segmentation approaches fine-tuned with RL treat the tool as fixed and already perfect~\cite{yun2026star, yao2025remotereasoner, zhang2024think2seg, liu2025seg, hu2025samr1, liu2025visionreasoner, du2026samveteran, zhu2025segagent}. c) JPTO enables optimization even if neither is true. Further, BTO allows for tool pretraining without rollout cost, while the two-stage bootstrapped method (GRTO) leads to improved policy convergence and the best final system.}
    \label{fig:relatedworks}
    \vspace{-3mm}
\end{figure}

\noindent\textbf{Multimodal LLMs for language-instructed segmentation.}
Early approaches to language-guided segmentation tightly couple a large language model with a pixel decoder via special segmentation tokens. LISA \cite{lai2023lisa} introduce this paradigm, appending a single \texttt{<SEG>} token to the LLM's vocabulary whose embedding is decoded by SAM \cite{kirillov2023sam} into a binary mask. Subsequent work extends LISA along several axes: LISA++ \cite{yang2023lisapp} and M2SA \cite{yuan2024m2sa} handle multiple objects per query, and SegLLM \cite{xu2024segllm} enable multi-turn conversational grounding. GLaMM \cite{rasheed2023glamm} enrich this design with grounded captions and region-level understanding, while GSVA \cite{xia2024gsva} incorporate rejection mechanisms for queries that lack a valid referent. X-SAM \cite{wang2024xsam} further generalizes the architecture toward open-vocabulary and versatile segmentation. Several designs beyond a \texttt{<SEG>} token are explored. RSVP \cite{hu2024rsvp} substitutes it with a BEiT-3-encoder \cite{beit3} over object tokens pre-detected by a MLLM. PixelLM \cite{ren2024pixellm} replaces it with a learned code-book of pixel embeddings for multi-target settings, and PSALM \cite{zhang2024psalm} adds rejection handling to it. A common limitation across this entire family is that all components are optimized with standard cross-entropy on fixed annotation sets, which can overfit to the label distribution, limiting generalization, and does not allow the model to explore novel approaches that could lead to superior results.

\noindent\textbf{Decoder as a tool and reinforcement learning.}
SAM4LLM \cite{chen2024sam4llm} is one of the first to train an MLLM to output bounding-boxes to prompt a frozen SAM2 \cite{ravi2024sam2}, using it as an external tool. While they used supervised fine-tuning (SFT), the success of group relative policy optimization (GRPO) \cite{shao2024deepseekmath} for reasoning in language models has inspired a second generation of segmentation approaches that use it to learn robust and accurate tool-use of SAM.

Seg-Zero \cite{liu2025seg} pioneer this regime, training the reasoning VLM with bounding-box IoU rewards, and using SAM itself only at inference. VisionReasoner \cite{liu2025visionreasoner} and POPEN \cite{zhu2025popen} extend Seg-Zero's multi-object handling, with POPEN using DPO \cite{rafailov2023dpo} on top of a LISA-pretrained backbone. SAM-R1 \cite{hu2025samr1} and StAR \cite{yun2026star} extend the approach to end-to-end rewards, evaluating SAM during training and using its mask IoU as reward. LENS \cite{zhu2025lens} and CoRPS \cite{lu2025coprs} suggest changing the prompting approach by adding a learnable context encoder and a correlation prior, respectively. To train introduced modules, they backpropagate a segmentation loss while tuning a MLLM with GRPO under mask-IoU rewards. Several attempts are made to employ these methods for specialized domains, e.g. RemoteReasoner \cite{yao2025remotereasoner} and Think2Seg \cite{zhang2024think2seg} in the remote sensing domain, or PathChat-SegR1\cite{liu2026pathchatsegr} in the pathology domain.

SAM3 agent \cite{carion2025sam3} suggests interacting with the segmentation decoder in a multi-turn, agentic manner. SAM-Veteran \cite{du2026samveteran} improves these interactions by multi-turn GRPO training, rewarding both mask and box quality across dialogue turns. RSAgent \cite{he2025rsagent} improves it further by adding agentic perception. SegAgent \cite{zhu2025segagent} explores a complementary agent formulation using Tree-of-Thoughts planning and a hybrid RL/SFT objective, enabling iterative mask refinement without gradient flow through SAM. These approaches demonstrate that RL-shaped language reasoning measurably lifts segmentation quality even when the segmentation tool itself is held fixed. However, most of the methods practically assume SAM to be a perfect tool and do not explore the potential gain in treating SAM as a trainable component.

\noindent\textbf{SAM fine-tuning for specialized domains.}
Although SAM exhibits strong zero-shot segmentation capabilities, its performance degrades on out-of-distribution imagery. Fine-tuning SAM on well-camouflaged animals, cast shadows, and medical images is shown to substantially recover accuracy in these failure modes \cite{chen2025sam3adapter}. Analogously, in remote sensing, fine-tuned SAM variants outperform prompt-tuned baselines on building footprints, solar panels, sinkholes, and dead-tree detection \cite{blushteinlivnon2026samfinetuningrs}. However, these studies assume perfect prompts, whereas in real-world scenarios prompts are often ambiguous or too long to be tokenized by SAM.

\noindent\textbf{Joint RL for dense outputs beyond segmentation.}
The composition of vision-language understanding and dense generation has been studied beyond segmentation. FireScope \cite{markov2025firescope} attaches a U-Net \cite{ronneberger2015unet} decoder to a reasoning model to generate wildfire risk maps, the MLLM and decoder are trained in stages with GRPO and regression loss respectively. Janus \cite{wu2024janus,chen2025januspro} interleave visual understanding and image generation within a single autoregressive transformer with a joint vocabulary. Several works \cite{duan2025got, pan2025januspror1advancingcollaborativevisual, ma2025stage} have applied GRPO to the Janus architecture for joint reasoning and image generation. Bagel \cite{deng2025bagel} similarly unite language tokens with normalization flows trained by flow matching for language-instructed image-to-image tasks. Several approaches \cite{liu2026unigrpounifiedpolicyoptimization, ye2026understandingvsgenerationnavigating} apply RL to Bagel with summed objectives of text GRPO and Flow-GRPO \cite{liu2025flowgrpo}. While these methods target generation rather than segmentation, they support the general finding that joint fine-tuning improves the quality of language-instructed dense outputs.

Motivated by these limitations, we study referring segmentation as a trainable tool-use problem in which both the reasoning policy and the segmentation decoder may need to adapt. Rather than treating the decoder as a fixed reward component or fine-tuning it independently under perfectly specified prompts, we propose GRTO: a two-stage framework that first bootstraps the tool toward rollouts expected to be reinforced, and then jointly optimizes the policy and tool within GRPO (Figure~\ref{fig:relatedworks}).

\section{Method}\label{sec:method}
\begin{figure}[t]
    \centering
    \includegraphics[width=\linewidth]{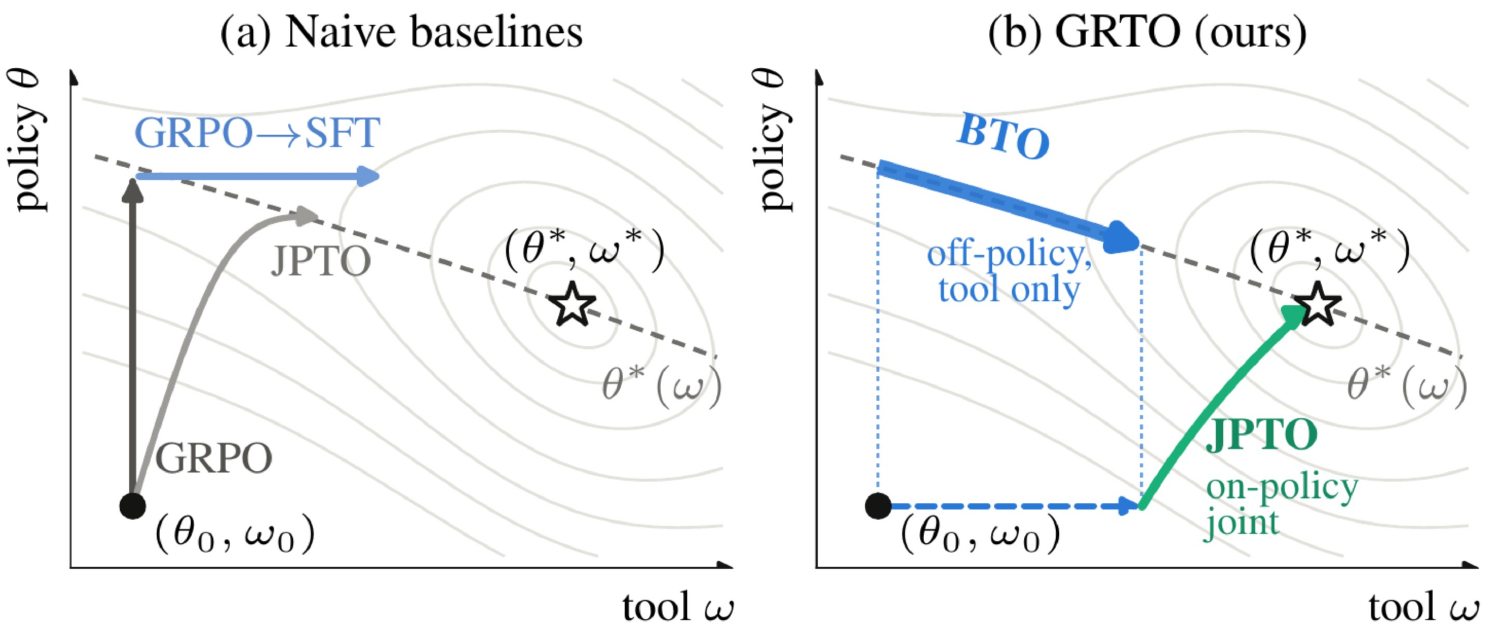}
    \caption{Training dynamics in the joint policy–tool weights space (schematic). GRPO and SFT climb only one axis, while JPTO alone start too far off. BTO fast-forwards the tool along the theoretically optimal policy $\pi_{\theta^*(\omega)}$, and only then fine-tunes with JPTO.}
    \label{fig:method}
    \vspace{-3mm}
\end{figure}

\noindent\textbf{Preliminary.}
We seek to optimize the standard KL-regularized RL objective \cite{korbak-etal-2022-rl,rafailov2023dpo,shao2024deepseekmath} for policy $\pi_{\theta}$, with a parametrized tool $\psi_{\omega}$ and reward $R$:
\begin{equation}\label{eq:klrl}
J_{\text{KL-RL}}(\theta,\omega) = \mathbb{E}_{\pi_\theta}[R(\psi_{\omega})] -\beta\,\text{D}_{\text{KL}}(\pi_{\theta}||\pi_{\theta_0}),
\end{equation}
where policy and tool are initialized from reference models $\pi_{\theta_0}$ and $\psi_{\omega_0}$. Commonly, in RL with tool-use one only trains the policy model, keeping the tool fixed, with updates:
\begin{equation}
    \theta_{0} \mapsto \argmax_{\theta} J_{\text{KL-RL}}(\theta,\omega_{0}), \quad \omega_0\mapsto\omega_0.
\end{equation}
To solve it for $\theta$, one employs Group Relative Policy Optimization (GRPO) \cite{deepseekai2025deepseekv32,liu2025understanding}, where a group of rollouts $\{o_1..o_G\}\sim\pi_{old}$ is sampled from the policy at the current optimization step, the mean $\mu$ of rewards $\{R_1..R_G\}$ is used as a baseline for the policy gradient, and their standard deviation $\sigma$ is used as normalization for numerical stability:

{\footnotesize
\begin{align}\label{eq:grpo}
J_{\text{GRPO}}(\theta,\omega_0)
    &= \frac{1}{G}\sum_{i=1}^{G}\sum_{t=1}^{|o_i|}
    \min\big(r_{i,t}A_i,\,
    \texttt{clip}(r_{i,t},1-\varepsilon,1+\varepsilon)A_i\big)
    \notag\\
    &\quad -\beta\,\text{D}_{\text{KL}}(\pi_\theta||\pi_{\theta_0}),
    \notag\\
\text{where,}\quad
    r_{i,t}
    &= \frac{\pi_\theta(o_{i,t}|o_{i,<t})}
    {\pi_{\text{old}}(o_{i,t}|o_{i,<t})},
    \quad
    A_i = \frac{R_i-\mu}{\sigma}.
\end{align}}
Here $\psi_{\omega_0}$ is treated as part of the reward model, and it is usually assumed that $\omega_0$ is perfect for the downstream task. While this assumption holds in some cases like code execution, it is not the case for complex parametrized tools derived from an optimization process of their own.


\noindent\textbf{Joint Optimization.}
When the tool $\psi_\omega$ is a pretrained model, it may under-perform under domain shifts \cite{blushteinlivnon2026samfinetuningrs,chen2025sam3adapter}. Moreover, one may treat $(\theta,\,\omega)$ as the single set of weights of the unified model, and in that case it is beneficial to allow them to align with each other during optimization. Therefore, our goal is to maximize a joint objective, with update:
\begin{equation}\label{eq:grto_goal}
\boxed{
    (\theta_0,\,\omega_0) \mapsto \argmax_{(\theta,\,\omega)} J_{\text{KL-RL}}(\theta,\omega).
    }
\end{equation}
For policy optimization, we use $J_{\text{GRPO}}$ \eqref{eq:grpo} with rewards calculated using the current tool parameters $\omega_{\text{old}}$. In other words, the policy, i.e. VLM, is updated using:
\begin{equation}
        \nabla_{\theta}J_{\text{KL-RL}} \approx \nabla_{\theta}J_{\text{GRPO}}(\theta,\omega_{\text{old}}).
\end{equation}
To estimate the tool gradient, we reuse the group of rollouts sampled from $\pi_{\text{old}}$ with sequence-level likelihood~\cite{zheng2025groupsequencepolicyoptimization}. We then \emph{approximate the reward gradient} with the negative gradient of a differentiable surrogate loss $\mathcal{L}$, providing a direct optimization signal for the tool:
\begin{equation}\label{eq:grto_grads}
\footnotesize
\begin{split}
    \nabla_{\omega}J_{\text{KL-RL}} = \mathbb{E}_{\pi_\theta}\nabla_{\omega} R(\psi_{\omega}) &\approx \mathbb{E}_{\pi_{\text{old}}}\left[
    \frac{\pi_\theta(o)}{\pi_{\text{old}}(o)}
    \nabla_{\omega}R(\psi_\omega)\right],
    \\
    &\approx -\frac{1}{G}\sum_{i=1}^{G}
    \left[\prod_{t=1}^{|o_i|}r_{i,t}\right]
    \nabla_{\omega}\mathcal{L}(\psi_\omega). 
\end{split}
\end{equation}
Thus, for the unified policy and tool optimization of \eqref{eq:grto_goal}, we define the joint policy-tool optimization (\textbf{JPTO}) objective:

\begin{equation}\label{eq:grto}
\boxed{
    J_{\text{JPTO}}(\theta,\omega)\!=\!J_{\text{GRPO}}(\theta,\omega_{\text{old}})\!-\!\frac{1}{G}\sum_{i=1}^{G}\left[\prod_{t=1}^{|o_i|}r'_{i,t}\right]\mathcal{L}(\psi_\omega).
    }
\end{equation}

Where $r'_{i,t}=\text{stopgrad}(r_{i,t})$ and the reward $R$ are treated as non-differentiable to reconstruct the gradients of~\eqref{eq:grto_grads}.

\noindent\textbf{Bootstrapping the Tool.}
A practical bottleneck in JPTO training is matching training velocities: $\psi_\omega$ is often a lightweight deterministic module that can be evaluated and updated cheaply, but during JPTO it can only be updated as fast as $\pi_\theta$ whose rollouts and updates are often vastly more expensive. This asymmetry suggests a natural bootstrapping strategy to first pretrain the tool using a fixed replay buffer of rollouts \cite{fedus2020replay}. A better-conditioned tool should also provide a more reliable reward signal during the second stage, thereby improving the stability and convergence of policy optimization. The challenge is formulating the pretraining objective to be consistent with $J_{\text{KL-RL}}$ \eqref{eq:klrl}.

To derive the consistent objective, we consider the optimal policy for any tool \(\psi_\omega\). For $\beta\!>\!0$, the policy that maximizes~\eqref{eq:klrl} has a closed-form Bayesian posterior~\(\pi_{\theta^*(\omega)}\)~\cite{korbak-etal-2022-rl}:
\begin{equation}
\pi_{\theta^*(\omega)} = \frac{1}{Z_\omega}\pi_{\theta_0}\,e^{\frac{1}{\beta}R(\psi_\omega)},\quad
Z_\omega=\mathbb{E}_{\pi_{\theta_0}} \left[e^{\frac{1}{\beta}R(\psi_\omega)}\right],
\end{equation}
where \(Z_\omega\) is the normalization constant making  $\pi_{\theta^*(\omega)}$ a normalized probability distribution. Intuitively, this posterior assigns higher probability to rollouts with higher rewards, while remaining are regularized toward the reference policy \(\pi_{\theta_0}\).
We can therefore rewrite the joint optimization as,
\begin{equation}
\begin{split}
\max_{\omega,\theta}\,J_{\text{KL-RL}}(\theta,\omega)
&= \max_{\omega}\,\max_{\theta}\,J_{\text{KL-RL}}(\theta,\omega)
\\
&= \max_{\omega}\,J_{\text{KL-RL}}(\theta^*(\omega),\omega),
\end{split}
\end{equation}
This shows that the optimization problem can be solved for tool parameters $\omega$ independently, while simultaneously aiming for $\theta^*$, as illustrated in Figure~\ref{fig:method}. In fact, substituting the posterior policy into the KL-regularized objective leads to the cancellation of the reward terms. The resulting objective depends only on the tool, which is derived below,
\begin{align}
\begin{split}
&J_{\text{KL-RL}}(\theta^*(\omega),\omega)
\\
&\quad= \mathbb{E}_{\pi_{\theta^*(\omega)}}
\left[ R(\psi_{\omega}) -\beta\,\log \left(\frac{\pi_{{\theta^*(\omega)}}}{\pi_{\theta_0}}\right) \right]
\\
&\quad= \mathbb{E}_{\pi_{\theta^*(\omega)}}
\left[R(\psi_{\omega}) -\beta\,\log \left( \frac{1}{Z_{\omega}}e^{\frac{1}{\beta}R(\psi_\omega)} \right) \right]
\\
&\quad= \mathbb{E}_{\pi_{\theta^*(\omega)}}[\beta\log Z_\omega] = \beta\log Z_\omega.
\end{split}
\end{align}
Thus, optimizing the original KL-regularized objective is equivalent to maximizing \(\beta\log Z_\omega\), for the optimal policy, allowing us to derive a tool-only optimization objective. Now, we are interested in its gradient w.r.t. the tool parameters $\omega$,
\begin{equation}\label{eq:bto_grads}
\begin{split}
\nabla_\omega\,J_{\text{KL-RL}}(\theta^*(\omega),\omega) &= \nabla_\omega\,\beta\log Z_{\omega} = \beta\frac{1}{Z_\omega}\nabla_{\omega}Z_{\omega}
\\
&\approx -\frac{1}{G}\sum_{i=1}^{G}\frac{1}{Z}e^{\frac{1}{\beta}R_i}\nabla_{\omega}\mathcal{L}(\psi_{\omega}),
\end{split}
\end{equation}
where we estimate the expectation using a group of \(G\) rollouts sampled from the reference policy, \(\{o_i\}_{i=1}^G\sim\pi_{\theta_0}\), and approximate the reward gradient using the surrogate loss \(\mathcal{L}\), as in \eqref{eq:grto_grads}. The exponential weights assign greater importance to rollouts with higher rewards. This yields our bootstrapped tool optimization (\textbf{BTO}) objective:
\begin{equation}\label{eq:bto}
\boxed{
\!J_{\text{BTO}}(\omega)\!\!=\!\!-\frac{1}{G}\!\sum_{i=1}^{G} \!
\frac{1}{Z}e^{\frac{1}{\beta}R_i}
\mathcal{L}(\psi_\omega;o_i),
\,\,\,
Z\!\!=\!\!\sum_{i=1}^G\!e^{\frac{1}{\beta}R_i}.
}
\end{equation}

Where \(Z\) and the rewards \(R_i\) are treated as non-differentiable. Noticing the striking similarity between BTO and the second term of the JPTO objective \eqref{eq:grto}, the gradients of BTO can be viewed as tool gradients in JPTO in an extreme off-policy setting, assuming $\theta\rightarrow\theta^*$ and $\theta_{old}\rightarrow\theta_0$, with importance sampling weights ${{\pi_{\theta^*}}/{\pi_{\theta_0}}}$. However, as this uses crude approximations of an idealized $\pi_{\theta^*}$, the bootstrapped tool would not perfectly align with the learned policy. Therefore, we still apply JPTO as a second stage to train $\theta$ and further fine-tune $\omega$.
Finally, we propose group relative tool optimization (\textbf{GRTO}) consisting of two stages:
\begin{equation}
\boxed{
(\theta_0,\omega_0) \mapsto (\theta_0, \argmax_{\omega} J_{\text{BTO}}(\omega)) \mapsto \argmax_{(\theta,\,\omega)} J_{\text{JPTO}}(\theta,\omega)
}
\end{equation}
where the preconditioned tool at the second stage improves convergence. For method dynamics illustration, see Figure~\ref{fig:method}.

\paragraph{GRTO in referring segmentation.} In referring segmentation, the policy $\pi_\theta$ is a VLM that reasons about the query and outputs prompts for the referred objects, and the tool $\psi_\omega$ is a promptable segmentation decoder that turns them into masks. The reward is the IoU of the final mask, and the tool is trained with a differentiable segmentation surrogate $\mathcal{L}$. We detail the models and prompting in Section~\ref{subsec:experimental_configurations}.
\section{Experiments}\label{sec:experimental_settings}

\begin{table*}[t]
    \caption{Main results. (a) Camouflage detection on COD10K~\cite{fan2021cod10k}, CAMO~\cite{ltnghia2019camo} and NC4K~\cite{yunqiu_cod21} (S: S-measure, wFm: weighted F-measure, E: mean E-measure, MAE: mean absolute error) (b) Remote sensing segmentation on EarthReason~\cite{li2025earthreason} (Rem.Reasoner: RemoteReasoner). (c) Reasoning segmentation on ReasonSeg~\cite{lai2023lisa}, ReasonSeg-R and ReasonSeg-X~\cite{yun2026star}, with baselines of similar backbone size (Q.: Qwen; MV: majority voting); StAR + GRPO/GRTO apply GRPO/GRTO to the StAR Qwen 3 checkpoint. The best result is in \textbf{bold} and top-2 results are \underline{underlined}.  $\dagger$: domain-specialized method. *: zero-shot evaluation (dataset not seen during training). Our results are two-seed means; the better of GRPO and GRTO is highlighted in gold.}
    \label{tab:main}
    \centering
    \setlength{\tabcolsep}{2pt}
    \sbox{\tabboxC}{%
    \begin{tabular}{l@{\hspace{3.5pt}}ccccccccccccc}
         \toprule
         \multirow{2}{*}{Method} & \multirow{2}{*}{$\dagger$} & \multicolumn{4}{c}{COD10K \cite{fan2021cod10k}} & \multicolumn{4}{c}{CAMO \cite{ltnghia2019camo}} & \multicolumn{4}{c}{NC4K* \cite{yunqiu_cod21}} \\
         & & S & wFm & E & MAE & S & wFm & E & MAE & S & wFm & E & MAE \\
         \midrule
         FEDER \cite{he2023camouflaged} & $\dagger$ & .823 & .740 & .911 & .032 & .807 & .785 & \underline{.947} & .069 & .846 & .817 & .916 & .045 \\
         ZoomNet \cite{pang2022zoom} & $\dagger$ & .830 & .729 & .911 & .029 & .820 & .752 & .892 & .066 & .853 & .784 & .912 & .043 \\
         SINet-v2 \cite{fan2021cod10k} & $\dagger$ & .815 & .680 & .906 & .037 & .820 & .743 & .895 & .070 & .847 & .805 & .914 & .048 \\
         HitNet \cite{hu2023high} & $\dagger$ & .868 & .798 & .932 & .024 & .844 & .801 & .902 & .057 & .870 & .825 & .921 & .039 \\
         OPNet \cite{mei2023camouflaged} & $\dagger$ & .857 & .767 & .919 & .026 & .858 & .817 & .915 & .050 & .883 & .838 & .932 & .034 \\
         DTINet \cite{liu2022boosting} & $\dagger$ & .824 & .695 & .896 & .034 & .856 & .796 & .916 & .050 & .863 & .792 & .917 & .041 \\
         BSA-Net \cite{zhu2022can} & $\dagger$ & .818 & .699 & .901 & .034 & .794 & .717 & .867 & .079 & .841 & .771 & .907 & .048 \\
         FocusDiffuser \cite{zhao2024focusdiffuser} & $\dagger$ & .875 & .809 & .939 & .020 & .881 & .851 & .939 & .042 & .891 & .854 & .940 & .029 \\
         FSEL \cite{sun2024frequency} & $\dagger$ & .873 & .800 & .928 & .021 & \underline{.885} & .851 & .942 & \underline{.040} & .892 & .853 & .941 & .030 \\
         C3Net \cite{jan2025c3net} & $\dagger$ & \underline{.898} & \underline{.851} & \underline{\textbf{.961}} & \underline{\textbf{.016}} & \underline{\textbf{.904}} & \underline{\textbf{.889}} & \underline{\textbf{.951}} & \underline{\textbf{.031}} & \underline{\textbf{.913}} & \underline{\textbf{.895}} & \underline{\textbf{.958}} & \underline{\textbf{.022}} \\
         \midrule
         GRPO & & .885 & .840 & .940 & .023 & .866 & .843 & .916 & .046 & .899 & .876 & .944 & .027 \\
         GRTO \textit{(ours)} & & \cellcolor{rankone}\underline{\textbf{.906}} & \cellcolor{rankone}\underline{\textbf{.874}} & \cellcolor{rankone}\underline{.958} & \cellcolor{rankone}\underline{\textbf{.016}} & \cellcolor{rankone}.878 & \cellcolor{rankone}\underline{.860} & \cellcolor{rankone}.930 & \cellcolor{rankone}.041 & \cellcolor{rankone}\underline{.910} & \cellcolor{rankone}\underline{\textbf{.895}} & \cellcolor{rankone}\underline{.953} & \cellcolor{rankone}\underline{.023} \\
         \bottomrule
    \end{tabular}
    }%
    \sbox{\tabboxA}{%
            \begin{tabular}{l@{\hspace{3.5pt}}ccccc}
        \toprule
         \multirow{3}{*}{Method} & \multirow{3}{*}{$\dagger$} & \multicolumn{4}{c}{EarthReason \cite{li2025earthreason}} \\
         & & \multicolumn{2}{c}{val} & \multicolumn{2}{c}{test} \\
         & & gIoU & cIoU & gIoU & cIoU \\
         \midrule
         Seg-Zero \cite{liu2025seg} & & 67.15 & 65.41 & 67.89 & 62.43 \\
         SegEarth-R1 \cite{li2025earthreason} & $\dagger$ & 68.25 & 64.13 & 70.75 & 68.60 \\
         Rem.Reasoner \cite{yao2025remotereasoner} & $\dagger$ & 69.13 & 67.80 & 70.96 & 69.02 \\
         GeoPixel \cite{shabbir2025geopixel} & $\dagger$ & 52.13 & 54.23 & 52.53 & 53.90 \\
         PSALM \cite{zhang2024psalm} & & 66.61 & 62.03 & 68.30 & 64.61 \\
         PixelLM \cite{ren2024pixellm} & & 57.94 & 57.79 & 60.01 & 59.22 \\ 
         LISA \cite{lai2023lisa} & & 59.10 & 57.39 & 60.88 & 61.04 \\
         UniGeoSeg \cite{ni2025unigeoseg} & $\dagger$ & \underline{72.54} & \underline{73.30} & 73.08 & 74.61 \\
         Think2Seg \cite{zhang2024think2seg} & & 72.16 & 72.87 & \underline{73.36} & \underline{75.60} \\
         \midrule
         GRPO & & \underline{73.91} & \underline{73.71} & \underline{74.26} & \underline{76.19} \\
         GRTO \textit{(ours)} & & \cellcolor{rankone}\underline{\textbf{76.01}} & \cellcolor{rankone}\underline{\textbf{77.25}} & \cellcolor{rankone}\underline{\textbf{76.49}} & \cellcolor{rankone}\underline{\textbf{78.95}} \\
         \bottomrule
    \end{tabular}
    }%
    \sbox{\tabboxB}{%
\begin{tabular}{l@{\hspace{3.5pt}}ccccccc}
        \toprule
         \multirow{3}{*}{Method} & \multirow{3}{*}{$\dagger$} & \multicolumn{2}{c}{RS* \cite{lai2023lisa}} & \multicolumn{2}{c}{RS-R* \cite{yun2026star}} & \multicolumn{2}{c}{RS-X \cite{yun2026star}} \\
         & & \multicolumn{2}{c}{test} & \multicolumn{2}{c}{-} & \multicolumn{2}{c}{test} \\
         & & gIoU & cIoU & gIoU & cIoU & gIoU & cIoU \\
         \midrule
         LISA \cite{lai2023lisa} & & 51.5 & 51.3 & 52.5 & 53.3 & 25.1 & 26.0 \\
         VisionReasoner \cite{liu2025visionreasoner} & & 63.6 & 55.7 & 64.8 & 56.8 & 42.2 & 33.8 \\
         SAM 3 Agent-7B \cite{carion2025sam3} & & 62.6 & 56.2 & 63.1 & 58.0 & 34.4 & 29.5 \\
         StAR Q.~2.5 (stage-1) \cite{yun2026star} & $\dagger$ & 66.7 & 60.8 & 69.0 & 65.6 & 47.4 & 40.6 \\
         StAR Q.~2.5 \cite{yun2026star} & $\dagger$ & 67.5 & 61.3 & 69.7 & 66.2 & 49.2 & 43.6 \\
         StAR Q.~2.5 + MV \cite{yun2026star} & $\dagger$ & 68.5 & 65.0 & 70.7 & 67.2 & 50.3 & 44.9 \\
         StAR Q.~3 \cite{yun2026star} & $\dagger$ & 70.6 & 63.8 & 73.8 & 69.2 & 57.9 & 50.1 \\
         StAR Q.~3 + MV \cite{yun2026star} & $\dagger$ & 71.8 & 66.5 & \underline{74.9} & \underline{69.7} & 59.6 & \underline{53.1} \\
         \midrule
         StAR + GRPO + MV & $\dagger$ & \underline{72.3} & \underline{66.9} & \underline{74.9} & 69.5 & \underline{59.9} & 52.9 \\
         StAR + GRTO + MV \textit{(ours)} & $\dagger$ & \cellcolor{rankone}\underline{\textbf{72.7}} & \cellcolor{rankone}\underline{\textbf{67.6}} & \cellcolor{rankone}\underline{\textbf{75.3}} & \cellcolor{rankone}\underline{\textbf{70.5}} & \cellcolor{rankone}\underline{\textbf{60.1}} & \cellcolor{rankone}\underline{\textbf{53.9}} \\
         \bottomrule
    \end{tabular}
    }%
    \pgfmathsetmacro{\tabRc}{\the\wd\tabboxC/(\the\ht\tabboxC+\the\dp\tabboxC)}%
    \pgfmathsetmacro{\tabRa}{\the\wd\tabboxA/(\the\ht\tabboxA+\the\dp\tabboxA)}%
    \pgfmathsetmacro{\tabRb}{\the\wd\tabboxB/(\the\ht\tabboxB+\the\dp\tabboxB)}%
    \pgfmathsetmacro{\tabH}{(\the\textwidth-5pt)/(\tabRc+\tabRa+\tabRb)}
    \begin{subtable}[t]{\tabRc\dimexpr\tabH pt\relax}%
        \centering
        \caption{Camouflage detection}
        \label{tab:camo}
        \resizebox*{!}{\tabH pt}{\usebox{\tabboxC}}%
    \end{subtable}\hspace{2pt}%
    \begin{subtable}[t]{\tabRa\dimexpr\tabH pt\relax}%
        \centering
        \caption{Remote sensing}
        \label{tab:earthreason}
        \resizebox*{!}{\tabH pt}{\usebox{\tabboxA}}%
    \end{subtable}\hspace{2pt}%
    \begin{subtable}[t]{\tabRb\dimexpr\tabH pt\relax}%
        \centering
        \caption{Reasoning segmentation}
        \label{tab:reasonseg}
        \resizebox*{!}{\tabH pt}{\usebox{\tabboxB}}%
    \end{subtable}%
\end{table*}

We conduct experiments on three challenging referring segmentation settings, which are presented separately below.

\noindent\textbf{Camouflage detection.} Camouflaged objects commonly exhibit much lower contrast and higher appearance similarity to their surroundings compared to common segmentation targets. COD10K~\cite{fan2021cod10k} and CAMO~\cite{ltnghia2019camo} are widely used camouflaged object segmentation benchmarks with pixel-level object mask annotations. As these datasets do not provide a separate validation split, a recent line of work has started using the testing set as validation set~\cite{chen2025sam3adapter, zheng2024birefnet, yadav2025samwave}. We instead follow the protocol of splitting the training set into train and validation subsets, and compare with the SoTA within that protocol~\cite{jan2025c3net}. In line with the common protocol, we report standard camouflaged object detection metrics: S-measure, weighted F-measure, E-measure, and mean absolute error; for comparability with the other settings, we additionally report gIoU and cIoU in the supplementary. We select checkpoints with the highest validation weighted F-measure. In line with previous works, we also benchmark final models zero-shot on a third camouflage detection dataset: NC4K~\cite{yunqiu_cod21}. Recent COD methods add task-specific priors to generic segmentation architectures, including boundary enhancement~\cite{he2023camouflaged,zhu2022can,jan2025c3net}, frequency or wavelet features~\cite{sun2024frequency,yadav2025samwave}, multi-scale context~\cite{pang2022zoom,mei2023camouflaged}, and foreground--background contrast or saliency disambiguation~\cite{jan2025c3net}. These designs reflect the need to recover subtle object boundaries and separate low-contrast camouflaged regions from visually similar backgrounds.

\noindent\textbf{Remote sensing segmentation.} Unlike natural images, remote sensing imagery typically captures expansive scenes with dense, repetitive structures and objects spanning widely varying scales. EarthReason~\cite{li2025earthreason} is a large-scale geospatial pixel reasoning benchmark comprising manually annotated image masks and implicit question-answer pairs for evaluating reasoning-guided remote sensing segmentation. We use it to train, test and evaluate in this setting. For checkpoint selection, we use a random subset of 200 validation questions, one per image, i.e., 2.9\% of the 6,802 validation questions. Since the remaining validation set is never used for selection, we additionally report results on the full validation set as a second, largely held-out test set. In line with current SoTA in this setting~\cite{zhang2024think2seg}, we report gIoU and cIoU: the average of per-sample IoUs, and the cumulative intersection over cumulative union.

Prior geospatial reasoning segmentation methods use remote-sensing-specific designs such as visual token compression, multi-scale language--vision fusion, spatial-correlation-aware mask generation, and task transformations for object-, region-, and pixel-level outputs~\cite{li2025earthreason,yao2025remotereasoner}. These adaptations address common properties of overhead imagery, including large spatial extent, small dense objects, and heterogeneous scene layouts.

\noindent\textbf{Challenging reasoning segmentation.} Challenging reasoning segmentation requires grounding targets from complex queries involving compositional, relational, contextual, or world-knowledge reasoning. ReasonSeg-X and ReasonSeg-R~\cite{yun2026star} are recently introduced benchmarks designed to stress-test reasoning-driven segmentation beyond standard referring expressions. ReasonSeg-R refines the original ReasonSeg benchmark~\cite{lai2023lisa}, providing test samples with corrected annotation and query issues. ReasonSeg-X extends the task with harder samples that require deeper and more systematic reasoning. Similarly to the current SoTA, we first pretrain the base model on common reasoning segmentation datasets before conducting the full range of experiments~\cite{yun2026star}. Following prior work, we report gIoU and cIoU. We select checkpoints on the ReasonSeg-X validation split; as results on a split used for model selection are optimistically biased, we report them only in the supplementary.

Recent SoTA methods for challenging reasoning segmentation invest additional computation directly into the reasoning process. In particular, StAR uses parallel test-time scaling through mask-level majority voting (MV) over multiple stochastic reasoning traces, effectively aggregating diverse candidate localizations before producing the final mask~\cite{yun2026star}. Its rollout-expanded selective tuning (REST) samples 128--256 trajectories but updates only on informative high- and low-advantage responses, mitigating weak signals when most rollouts are incorrect or uninformative.

For the complete prompts across all experiments, see the supplementary material.

\subsection{Experimental configurations}\label{subsec:experimental_configurations}

\noindent\textbf{Models and training configurations.} To compare GRTO to domain-specific methods, we select a general-purpose VLM: InternVL3.5-8B as our VLM backbone across camouflage and remote sensing experiments due to its strong performance across various domains and tasks including visual grounding~\cite{wang2025internvl35}, and SAM3 as our mask generator~\cite{carion2025sam3}. The policy generates a concise noun phrase and bounding boxes for the referred object instances, which prompt SAM3 in promptable concept segmentation mode. We then apply deterministic spatial filtering, setting all pixels outside the generated boxes to background, so that the boxes serve both as prompts and as hard spatial constraints. However, GRTO does not have to compete with domain-specific methods: they can be combined to achieve the best performance. To demonstrate this complementarity where the computational gap is largest, we focus on reasoning segmentation, where domain-specific methods require far more compute than standard approaches. We take the current SoTA -- StAR~\cite{yun2026star} -- and post-train their Qwen3-VL-8B~\cite{bai2025qwen3vl} checkpoint, applying GRTO with their REST training strategy during training. Following StAR, the policy reasons about the query before generating a bounding box and a point for each object, which prompt the mask decoder of SAM2.1~\cite{ravi2024sam2}. We then employ their MV strategy at inference. Across all experiments, we optimize SAM using a differentiable BCE + soft-IoU surrogate, while evaluating the VLM with final-mask IoU as the reward signal. We perform every experiment two times with different seeds and report means; standard deviations are given in the supplementary. For the detailed experimental setup, see the supplementary.

\noindent\textbf{Main experiments.} For each of the three settings, we train two configuration: standard \textbf{GRPO}, where we only fine-tune the base model, treating SAM as a frozen auxiliary tool; and full \textbf{GRTO}, where we pretrain SAM by pre-generating groups of rollouts with the reference policy once and backpropagating the objective in \eqref{eq:bto} on SAM until validation performance peaks, and proceed with standard JPTO training afterwards. In GRTO bootstrapping, we use experience replay~\cite{fedus2020replay}: rollout groups are generated once, stored, and treated as a static fine-tuning dataset. Its overhead is comparable to one GRPO epoch over the full dataset, since after groups are rolled out once, the computational cost of fine-tuning SAM is comparatively negligible.

\noindent\textbf{Ablations.} GRTO has two stages, BTO and JPTO. We ablate each stage with a group of three configurations and compare all of them to GRTO. Here, SFT denotes standard fine-tuning of SAM on policy rollouts, without BTO's reweighting in $J_{\text{BTO}}$ \eqref{eq:bto}, using as many rollouts as the BTO replay buffer. \emph{Is joint optimization necessary?} We train the tool but separately from the policy, which is trained with GRPO. In \textbf{GRPO$\rightarrow$SFT}, SAM is fine-tuned with SFT on the rollouts of the trained policy. In \textbf{SFT$\rightarrow$GRPO} and \textbf{BTO$\rightarrow$GRPO}, SAM is first trained on the replay buffer with SFT or BTO, respectively, and kept frozen during GRPO. BTO$\rightarrow$GRPO is GRTO without joint optimization; it remains useful in memory-bound settings where the policy and tool cannot be trained jointly. \emph{Is bootstrapping necessary?} We keep JPTO and change the tool training strategy. \textbf{JPTO} starts from the frozen SAM checkpoint without bootstrapping. \textbf{SFT$\rightarrow$JPTO} replaces BTO with SFT on the same replay buffer. \textbf{JPTO$\rightarrow$SFT} trains the tool with SFT after JPTO instead of before it, on the rollouts of the trained policy.

\noindent\textbf{Analysis.} We analyze how GRTO's gains relate to the frozen decoder's fit to each domain, and where they concentrate. We first quantify the \emph{domain shift}: how poorly the frozen SAM's pretrained features separate a domain's targets from their background. We measure it with the Gaussian Bhattacharyya coefficient (GBC)~\cite{pandy2022transferability} between target and background pixels in the prompt-free SAM3 image features, averaged over all images of each test set and then over the test sets; a lower GBC indicates a larger gap. To analyze where GRTO improves over GRPO, we compare the gain on all samples with the gain on two subsets fixed per test set: \emph{small} targets (the 20\% smallest ground-truth masks relative to the image) and \emph{hard} samples. To select hard samples independently of the compared models, we evaluate the frozen SAM with pseudo-ground-truth prompts derived from the annotations (the box of the ground-truth mask, with the class name where available; see the supplementary)~\cite{kirillov2023sam,ren2024segment} and take the 20\% of targets on which it scores lowest. We restrict this to single-region targets, since a single ground-truth box cannot specify a target that consists of several disconnected regions.

\subsection{Results} \label{subsec:results}

\begin{figure}[t]
    \centering
    \captionsetup[subfigure]{justification=centering}
    \begin{subfigure}[t]{0.535\linewidth}
        \centering
        \includegraphics[width=\linewidth]{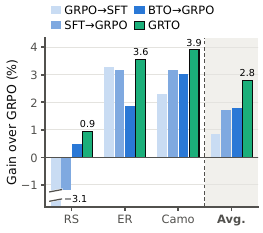}
        \caption{Is joint optimization necessary?}
        \label{fig:abl_joint}
    \end{subfigure}\hfill
    \begin{subfigure}[t]{0.459\linewidth}
        \centering
        \includegraphics[width=\linewidth]{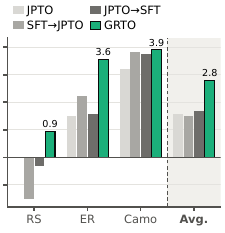}
        \caption{Is bootstrapping necessary?}
        \label{fig:abl_bootstrap}
    \end{subfigure}
    \caption{Ablations: relative gain over GRPO (mean of gIoU and cIoU; two-seed means over each domain's test sets). (a) The tool is trained separately from the policy, before (SFT$\rightarrow$GRPO, BTO$\rightarrow$GRPO) or after (GRPO$\rightarrow$SFT) GRPO. (b) JPTO without bootstrapping (JPTO), after SFT instead of BTO (SFT$\rightarrow$JPTO), or followed by SFT (JPTO$\rightarrow$SFT). RS: ReasonSeg; ER: EarthReason; Camo: camouflage; Avg.: mean over the three domains. Full results are in the supplementary material.}
    \label{fig:ablations}
\end{figure}

\begin{figure}[t]
    \centering
    \captionsetup[subfigure]{justification=centering}
    \begin{subfigure}[t]{0.49\linewidth}
        \centering
        \includegraphics[width=\linewidth]{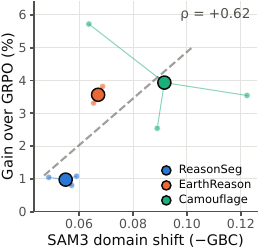}
        \caption{Gain vs.\ domain shift}
        \label{fig:transfer_corr}
    \end{subfigure}\hfill
    \begin{subfigure}[t]{0.49\linewidth}
        \centering
        \includegraphics[width=\linewidth]{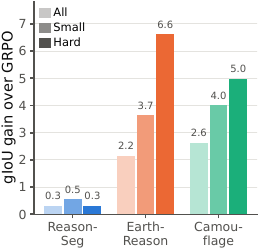}
        \caption{Gain on small and hard samples}
        \label{fig:concentrated}
    \end{subfigure}
    \caption{Where GRTO's gains arise. (a) GRTO's relative gain over GRPO (mean of gIoU and cIoU) vs.\ the SAM3 domain shift. Small markers: test sets; large markers: domain means; dashed: fit through the means; $\rho$: Spearman correlation over the test sets. (b) gIoU gain over GRPO on all samples, small targets, and hard samples (Section~\ref{subsec:experimental_configurations}).}
    \label{fig:analysis}
    \vspace{-5mm}
\end{figure}

Table~\ref{tab:main} reports the main results, Figure~\ref{fig:ablations} the ablations, and Figure~\ref{fig:analysis} the analysis of where GRTO's gains arise.

\noindent\textbf{Camouflage detection.} GRTO improves over GRPO on every metric of all three test sets (Table~\ref{tab:camo}). On COD10K, it sets a new state of the art in S-measure (.906) and weighted F-measure (.874, $+$.034 over GRPO), ties the best MAE (.016), and is second only to C3Net in E-measure. On the zero-shot NC4K, it ranks first or second on every metric, matching the best weighted F-measure (.895). On CAMO, it ranks in the top five on all metrics.

\noindent\textbf{Remote sensing segmentation.} GRTO achieves the best results on both EarthReason splits (Table~\ref{tab:earthreason}), with 76.49 gIoU and 78.95 cIoU on the test set. This improves over GRPO by $+$2.2 gIoU and $+$2.8 cIoU, and over the previous state of the art~\cite{zhang2024think2seg} by $+$3.1 gIoU and $+$3.4 cIoU. On the validation set, the gains over GRPO reach $+$2.1 gIoU and $+$3.5 cIoU.

\noindent\textbf{Challenging reasoning segmentation.} Applied to StAR, GRTO achieves the best gIoU and cIoU on all three benchmarks (Table~\ref{tab:reasonseg}), including the zero-shot ReasonSeg and ReasonSeg-R. It improves over StAR Qwen 3 with majority voting, the previous state of the art, by up to $+$0.9 gIoU and $+$1.1 cIoU, and over GRPO by up to $+$0.4 gIoU and $+$1.0 cIoU.

\noindent\textbf{Significance.} We test GRTO against GRPO with a paired bootstrap~\cite{efron1993introduction} over evaluation images (10{,}000 resamples, averaged over seeds and test sets), with one comparison per domain (mean of gIoU and cIoU for ReasonSeg and EarthReason, $F^w_\beta$ for camouflage) and Holm correction~\cite{holm1979simple}. The improvement is significant on EarthReason ($+2.63$, 95\% CI $[+2.03, +3.25]$) and camouflage ($F^w_\beta$ $+2.31$ $[+1.81, +2.82]$, both $p<0.001$), positive in each seed and on every test set. On ReasonSeg, it is smaller and marginally significant ($+0.63$ $[-0.00, +1.24]$ on the mean of gIoU and cIoU, $p=0.05$), consistent with its small domain shift (Figure~\ref{fig:transfer_corr}). Details are in the supplementary.

\noindent\textbf{Ablations.} GRTO outperforms every ablation in every domain (Figure~\ref{fig:ablations}). Averaged over the domains, it improves over GRPO by 2.8\%, while no ablation exceeds 1.8\%. \emph{Joint optimization.} Training the tool separately from the policy helps on EarthReason and camouflage, but GRPO$\rightarrow$SFT and SFT$\rightarrow$GRPO fall below GRPO on ReasonSeg ($-$3.1\% and $-$1.2\%). BTO$\rightarrow$GRPO is the only separate variant that improves in all three domains, but it gains less than GRTO. \emph{Bootstrapping.} Without BTO, joint optimization alone (JPTO) does not improve ReasonSeg and reaches less than half of GRTO's gain on EarthReason (1.5\% vs.\ 3.6\%). Neither replacing BTO with SFT nor applying SFT after JPTO closes this gap: both stay below GRTO in every domain and below GRPO on ReasonSeg.

\noindent\textbf{Analysis.} GRTO's gains grow with the domain shift of the frozen SAM (Figure~\ref{fig:transfer_corr}). Ordered by the extent of the domain shift of SAM3, the domains show increasing relative gains over GRPO: 1.0\% on ReasonSeg, 3.6\% on EarthReason, and 3.9\% on camouflage. Across the eight test sets, the gain correlates with the domain shift (Spearman's $\rho = 0.62$). Moreover, GRTO's gains concentrate on the most challenging samples (Figure~\ref{fig:concentrated}). On EarthReason, the gIoU gain over GRPO rises from $+$2.2 on all samples to $+$3.7 on small targets and $+$6.6 on hard samples; on camouflage, from $+$2.6 to $+$4.0 and $+$5.0, respectively.

\section{Discussion}

\noindent\textbf{GRTO as a general-purpose, composable method.}
Without domain-specific architectural priors, GRTO sets a new state of the art on EarthReason and COD10K and is competitive on CAMO and NC4K with methods tailored to each domain (Table~\ref{tab:main}). It is also complementary to such methods: applied on top of StAR, it improves the state of the art on all three reasoning segmentation benchmarks. GRTO can therefore serve both as a standalone approach that pairs MLLM reasoning with an adaptable tool, and as a plug-in that strengthens specialized pipelines.

\noindent\textbf{Both stages of GRTO are necessary.}
The two stages play complementary roles. BTO aligns the tool offline with an approximation of the trained policy, and JPTO corrects the remaining mismatch online, as the policy actually evolves. Without joint optimization, the tool is fit to an idealized policy that may not perfectly match the post-trained one. Without bootstrapping, joint optimization starts from a tool that transfers poorly, so policy and tool must adapt to each other from a weak starting point. Only their combination gives the best results in every domain.

\noindent\textbf{Naive fine-tuning can degrade the decoder.}
SFT weights all rollouts of a group equally, so the decoder also fits the masks of poor rollouts. This is most harmful on ReasonSeg, where the policy is challenged most: following StAR~\cite{yun2026star}, each group contains 64 rollouts, increasing the number of degenerate prompts. On ReasonSeg, every configuration with SFT falls below its counterpart without it and degrades the GRPO baseline. BTO's reweighting strategy avoids this \eqref{eq:bto}, and only the configurations with BTO improve over GRPO on ReasonSeg.

\paragraph{Bootstrapping is a cheap and effective way to align the tool.}
Bootstrapping alone recovers a significant amount of GRTO's gain. BTO$\rightarrow$GRPO, which differs from GRPO only in the tool's initialization, attains 64\% of GRTO's improvement on over GRPO on average (Figure~\ref{fig:ablations}). Because BTO only fine-tunes the tool on a static replay buffer, its cost is comparable to a single rollout epoch, making it a practical alternative when joint optimization is not affordable. Followed by JPTO, it yields the best results.

\noindent\textbf{Limitations.}
The bootstrapping stage depends on the quality and diversity of the replay buffer. If the reference policy almost never generates useful prompts for a class of examples, BTO has little signal for those cases. Larger groups and higher sampling variance mitigate this, at the cost of more rollouts and storage. We analyze the replay buffer and the KL divergence between the reference and trained policies in the supplementary material.

\noindent\textbf{Future work.}
We deliberately use a simple static replay buffer, yet the bootstrapped tool already improves results. More adaptive bootstrapping, such as buffers refreshed as the policy improves, curricula for difficult or underrepresented examples, or buffers mixing reference- and intermediate-policy rollouts, may yield further gains, suggesting that investing additional computation in tool bootstrapping is a practical lever for final performance. Finally, while we study referring segmentation, GRTO applies to any setting in which a policy calls a differentiable, trainable tool, such as detection, retrieval, or depth and pose estimation.

\section{Conclusion}

We present GRTO, a framework for optimizing a reasoning policy together with the differentiable tool it calls. GRTO first bootstraps the tool with BTO, which fine-tunes it on reference-policy rollouts weighted toward the optimal post-trained policy, and then optimizes policy and tool jointly with JPTO, reusing the same GRPO rollouts for policy updates and importance-weighted tool updates.

Across camouflage detection, remote sensing segmentation, and challenging reasoning segmentation, GRTO improves over policy-only GRPO and over every ablated configuration. Without domain-specific architectures, it sets a new state of the art on EarthReason and COD10K, and applied on top of StAR, it improves the state of the art on all three reasoning segmentation benchmarks. Its gains grow with the domain shift of the frozen tool and are largest on small and hard samples.

These results show that trainable tools should not be treated as fixed reward components. Aligning the tool with the policy objective, starting with a cheap bootstrapping stage, makes GRTO a general strategy for combining MLLM reasoning with adaptable differentiable tools.


{
    \small
    \bibliographystyle{ieeenat_fullname}
    \bibliography{main}
}

\clearpage
\appendix

\section{Full results}\label{sec:full_results}

Table~\ref{tab:full} reports GRTO and GRPO on every evaluated split with the standard deviation over the two seeds, including gIoU and cIoU for camouflage detection and the ReasonSeg-X validation split, which is used for checkpoint selection.


\begin{table*}[t]
\centering
\caption{Full results of GRTO and GRPO on every evaluated split, including gIoU and cIoU for camouflage detection and the ReasonSeg-X validation split used for checkpoint selection ($\dagger$). Reasoning segmentation: StAR + majority voting. Mean $\pm$ standard deviation over two seeds; best mean per column in \textbf{bold}. *: zero-shot. RS: ReasonSeg; RS-X: ReasonSeg-X; RS-R: ReasonSeg-R.}
\label{tab:full}
\setlength{\tabcolsep}{2.5pt}
\sbox{\tabboxA}{%
\begin{tabular}{l|cccc|cccc|cccc}
\toprule
\multirow{2}{*}{Method} & \multicolumn{4}{c|}{COD10K} & \multicolumn{4}{c|}{CAMO} & \multicolumn{4}{c}{NC4K*} \\
 & S & wFm & E & MAE$\downarrow$ & S & wFm & E & MAE$\downarrow$ & S & wFm & E & MAE$\downarrow$ \\
\midrule
GRPO & .885\,{\scriptsize$\pm$.0002} & .840\,{\scriptsize$\pm$.0003} & .940\,{\scriptsize$\pm$.0003} & .023\,{\scriptsize$\pm$.0002} & .866\,{\scriptsize$\pm$.0002} & .843\,{\scriptsize$\pm$.0003} & .916\,{\scriptsize$\pm$.0002} & .046\,{\scriptsize$\pm$.0002} & .899\,{\scriptsize$\pm$.0003} & .876\,{\scriptsize$\pm$.0003} & .944\,{\scriptsize$\pm$.0002} & .027\,{\scriptsize$\pm$.0002} \\
GRTO \textit{(ours)} & \textbf{.906}\,{\scriptsize$\pm$.0003} & \textbf{.874}\,{\scriptsize$\pm$.0002} & \textbf{.958}\,{\scriptsize$\pm$.0004} & \textbf{.016}\,{\scriptsize$\pm$.0003} & \textbf{.878}\,{\scriptsize$\pm$.0024} & \textbf{.860}\,{\scriptsize$\pm$.0057} & \textbf{.930}\,{\scriptsize$\pm$.0029} & \textbf{.041}\,{\scriptsize$\pm$.0018} & \textbf{.910}\,{\scriptsize$\pm$.0002} & \textbf{.895}\,{\scriptsize$\pm$.0001} & \textbf{.953}\,{\scriptsize$\pm$.0005} & \textbf{.023}\,{\scriptsize$\pm$.0004} \\
\bottomrule
\end{tabular}
}%
\begin{subtable}[t]{\textwidth}%
\centering
\caption{Camouflage detection}
\label{tab:full_camo}
\resizebox{\ifdim\wd\tabboxA>\textwidth\textwidth\else\wd\tabboxA\fi}{!}{\usebox{\tabboxA}}%
\end{subtable}
\par\medskip
\sbox{\tabboxA}{%
\begin{tabular}{l|cc|cc|cc}
\toprule
\multirow{2}{*}{Method} & \multicolumn{2}{c|}{COD10K} & \multicolumn{2}{c|}{CAMO} & \multicolumn{2}{c}{NC4K*} \\
 & gIoU & cIoU & gIoU & cIoU & gIoU & cIoU \\
\midrule
GRPO & 78.09\,{\scriptsize$\pm$0.02} & 78.88\,{\scriptsize$\pm$0.04} & 78.83\,{\scriptsize$\pm$0.03} & 73.74\,{\scriptsize$\pm$0.03} & 82.41\,{\scriptsize$\pm$0.03} & 83.99\,{\scriptsize$\pm$0.03} \\
GRTO \textit{(ours)} & \textbf{81.87}\,{\scriptsize$\pm$0.02} & \textbf{84.09}\,{\scriptsize$\pm$0.34} & \textbf{80.72}\,{\scriptsize$\pm$0.39} & \textbf{77.20}\,{\scriptsize$\pm$1.51} & \textbf{84.61}\,{\scriptsize$\pm$0.04} & \textbf{86.02}\,{\scriptsize$\pm$0.14} \\
\bottomrule
\end{tabular}
}%
\sbox{\tabboxB}{%
\begin{tabular}{l|cc|cc}
\toprule
\multirow{2}{*}{Method} & \multicolumn{2}{c|}{Val} & \multicolumn{2}{c}{Test} \\
 & gIoU & cIoU & gIoU & cIoU \\
\midrule
GRPO & 73.91\,{\scriptsize$\pm$0.65} & 73.71\,{\scriptsize$\pm$0.37} & 74.26\,{\scriptsize$\pm$0.67} & 76.19\,{\scriptsize$\pm$0.39} \\
GRTO \textit{(ours)} & \textbf{76.01}\,{\scriptsize$\pm$0.34} & \textbf{77.25}\,{\scriptsize$\pm$0.38} & \textbf{76.49}\,{\scriptsize$\pm$0.35} & \textbf{78.95}\,{\scriptsize$\pm$0.71} \\
\bottomrule
\end{tabular}
}%
\pgfmathsetmacro{\tabRa}{\the\wd\tabboxA/(\the\ht\tabboxA+\the\dp\tabboxA)}%
\pgfmathsetmacro{\tabRb}{\the\wd\tabboxB/(\the\ht\tabboxB+\the\dp\tabboxB)}%
\pgfmathsetmacro{\tabH}{(\the\textwidth-3pt)/(\tabRa+\tabRb)}%
\pgfmathsetmacro{\tabSprev}{\tabH/(\the\ht\tabboxA+\the\dp\tabboxA)}%
\begin{subtable}[t]{\tabRa\dimexpr\tabH pt\relax}%
\centering
\caption{Camouflage detection, gIoU and cIoU}
\label{tab:full_camo_iou}
\resizebox*{!}{\tabH pt}{\usebox{\tabboxA}}%
\end{subtable}\hspace{2pt}%
\begin{subtable}[t]{\tabRb\dimexpr\tabH pt\relax}%
\centering
\caption{Remote sensing}
\label{tab:full_er}
\resizebox*{!}{\tabH pt}{\usebox{\tabboxB}}%
\end{subtable}%
\par\medskip
\sbox{\tabboxA}{%
\begin{tabular}{l|cc|cc|cc|cc}
\toprule
\multirow{2}{*}{Method} & \multicolumn{2}{c|}{RS-X val$^\dagger$} & \multicolumn{2}{c|}{RS-X test} & \multicolumn{2}{c|}{RS-R*} & \multicolumn{2}{c}{RS*} \\
 & gIoU & cIoU & gIoU & cIoU & gIoU & cIoU & gIoU & cIoU \\
\midrule
GRPO & \textbf{63.3}\,{\scriptsize$\pm$1.0} & \textbf{61.1}\,{\scriptsize$\pm$2.6} & 59.9\,{\scriptsize$\pm$0.6} & 52.9\,{\scriptsize$\pm$1.0} & 74.9\,{\scriptsize$\pm$0.1} & 69.5\,{\scriptsize$\pm$0.2} & 72.3\,{\scriptsize$\pm$0.2} & 66.9\,{\scriptsize$\pm$1.2} \\
GRTO \textit{(ours)} & 62.0\,{\scriptsize$\pm$0.5} & 56.9\,{\scriptsize$\pm$1.7} & \textbf{60.1}\,{\scriptsize$\pm$0.8} & \textbf{53.9}\,{\scriptsize$\pm$1.0} & \textbf{75.3}\,{\scriptsize$\pm$0.1} & \textbf{70.5}\,{\scriptsize$\pm$0.8} & \textbf{72.7}\,{\scriptsize$\pm$0.4} & \textbf{67.6}\,{\scriptsize$\pm$0.5} \\
\bottomrule
\end{tabular}
}%
\begin{subtable}[t]{\textwidth}%
\centering
\caption{Reasoning segmentation}
\label{tab:full_rs}
\scalebox{\tabSprev}{\usebox{\tabboxA}}%
\end{subtable}
\end{table*}

\begin{table*}[t]
\centering
\caption{Ablations on every evaluated split, including the ReasonSeg-X validation split used for checkpoint selection ($\dagger$). Rows: GRPO; the ablations without joint optimization (GRPO$\rightarrow$SFT, SFT$\rightarrow$GRPO, BTO$\rightarrow$GRPO); the ablations without bootstrapping (JPTO, SFT$\rightarrow$JPTO, JPTO$\rightarrow$SFT); GRTO. Reasoning segmentation: StAR + majority voting. Mean $\pm$ standard deviation over two seeds; best mean per column in \textbf{bold}. *: zero-shot. RS: ReasonSeg; RS-X: ReasonSeg-X; RS-R: ReasonSeg-R.}
\label{tab:abl}
\setlength{\tabcolsep}{2.5pt}
\sbox{\tabboxA}{%
\begin{tabular}{l|cccc|cccc|cccc}
\toprule
\multirow{2}{*}{Method} & \multicolumn{4}{c|}{COD10K} & \multicolumn{4}{c|}{CAMO} & \multicolumn{4}{c}{NC4K*} \\
 & S & wFm & E & MAE$\downarrow$ & S & wFm & E & MAE$\downarrow$ & S & wFm & E & MAE$\downarrow$ \\
\midrule
GRPO & .885\,{\scriptsize$\pm$.0002} & .840\,{\scriptsize$\pm$.0003} & .940\,{\scriptsize$\pm$.0003} & .023\,{\scriptsize$\pm$.0002} & .866\,{\scriptsize$\pm$.0002} & .843\,{\scriptsize$\pm$.0003} & .916\,{\scriptsize$\pm$.0002} & .046\,{\scriptsize$\pm$.0002} & .899\,{\scriptsize$\pm$.0003} & .876\,{\scriptsize$\pm$.0003} & .944\,{\scriptsize$\pm$.0002} & .027\,{\scriptsize$\pm$.0002} \\
\midrule
GRPO$\rightarrow$SFT & .899\,{\scriptsize$\pm$.0054} & .867\,{\scriptsize$\pm$.0057} & .951\,{\scriptsize$\pm$.0052} & .019\,{\scriptsize$\pm$.0018} & .868\,{\scriptsize$\pm$.0044} & .848\,{\scriptsize$\pm$.0050} & .919\,{\scriptsize$\pm$.0044} & .045\,{\scriptsize$\pm$.0023} & .904\,{\scriptsize$\pm$.0034} & .888\,{\scriptsize$\pm$.0036} & .947\,{\scriptsize$\pm$.0038} & .025\,{\scriptsize$\pm$.0010} \\
SFT$\rightarrow$GRPO & .904\,{\scriptsize$\pm$.0010} & .869\,{\scriptsize$\pm$.0011} & .955\,{\scriptsize$\pm$.0000} & .017\,{\scriptsize$\pm$.0002} & .869\,{\scriptsize$\pm$.0011} & .850\,{\scriptsize$\pm$.0023} & .922\,{\scriptsize$\pm$.0009} & .047\,{\scriptsize$\pm$.0003} & .909\,{\scriptsize$\pm$.0006} & .894\,{\scriptsize$\pm$.0012} & \textbf{.953}\,{\scriptsize$\pm$.0006} & .024\,{\scriptsize$\pm$.0003} \\
BTO$\rightarrow$GRPO & .902\,{\scriptsize$\pm$.0005} & .867\,{\scriptsize$\pm$.0006} & .953\,{\scriptsize$\pm$.0004} & .018\,{\scriptsize$\pm$.0005} & .875\,{\scriptsize$\pm$.0067} & .858\,{\scriptsize$\pm$.0085} & .929\,{\scriptsize$\pm$.0102} & \textbf{.041}\,{\scriptsize$\pm$.0035} & .908\,{\scriptsize$\pm$.0001} & .891\,{\scriptsize$\pm$.0008} & .951\,{\scriptsize$\pm$.0001} & .024\,{\scriptsize$\pm$.0000} \\
\midrule
JPTO & .905\,{\scriptsize$\pm$.0009} & .873\,{\scriptsize$\pm$.0005} & .956\,{\scriptsize$\pm$.0007} & .017\,{\scriptsize$\pm$.0002} & .871\,{\scriptsize$\pm$.0007} & .853\,{\scriptsize$\pm$.0024} & .926\,{\scriptsize$\pm$.0008} & .044\,{\scriptsize$\pm$.0005} & \textbf{.910}\,{\scriptsize$\pm$.0007} & .895\,{\scriptsize$\pm$.0018} & .952\,{\scriptsize$\pm$.0006} & .024\,{\scriptsize$\pm$.0002} \\
SFT$\rightarrow$JPTO & \textbf{.908}\,{\scriptsize$\pm$.0007} & \textbf{.878}\,{\scriptsize$\pm$.0004} & \textbf{.958}\,{\scriptsize$\pm$.0005} & \textbf{.016}\,{\scriptsize$\pm$.0002} & .874\,{\scriptsize$\pm$.0012} & .857\,{\scriptsize$\pm$.0010} & .927\,{\scriptsize$\pm$.0006} & .043\,{\scriptsize$\pm$.0003} & \textbf{.910}\,{\scriptsize$\pm$.0010} & \textbf{.896}\,{\scriptsize$\pm$.0008} & .952\,{\scriptsize$\pm$.0007} & .024\,{\scriptsize$\pm$.0005} \\
JPTO$\rightarrow$SFT & .906\,{\scriptsize$\pm$.0003} & .876\,{\scriptsize$\pm$.0017} & .957\,{\scriptsize$\pm$.0000} & \textbf{.016}\,{\scriptsize$\pm$.0000} & .873\,{\scriptsize$\pm$.0018} & .856\,{\scriptsize$\pm$.0019} & .928\,{\scriptsize$\pm$.0001} & .043\,{\scriptsize$\pm$.0002} & .909\,{\scriptsize$\pm$.0012} & .895\,{\scriptsize$\pm$.0014} & .952\,{\scriptsize$\pm$.0004} & .024\,{\scriptsize$\pm$.0004} \\
\midrule
GRTO \textit{(ours)} & .906\,{\scriptsize$\pm$.0003} & .874\,{\scriptsize$\pm$.0002} & \textbf{.958}\,{\scriptsize$\pm$.0004} & \textbf{.016}\,{\scriptsize$\pm$.0003} & \textbf{.878}\,{\scriptsize$\pm$.0024} & \textbf{.860}\,{\scriptsize$\pm$.0057} & \textbf{.930}\,{\scriptsize$\pm$.0029} & \textbf{.041}\,{\scriptsize$\pm$.0018} & \textbf{.910}\,{\scriptsize$\pm$.0002} & .895\,{\scriptsize$\pm$.0001} & \textbf{.953}\,{\scriptsize$\pm$.0005} & \textbf{.023}\,{\scriptsize$\pm$.0004} \\
\bottomrule
\end{tabular}
}%
\begin{subtable}[t]{\textwidth}%
\centering
\caption{Camouflage detection}
\label{tab:abl_camo}
\resizebox{\ifdim\wd\tabboxA>\textwidth\textwidth\else\wd\tabboxA\fi}{!}{\usebox{\tabboxA}}%
\end{subtable}
\par\medskip
\sbox{\tabboxA}{%
\begin{tabular}{l|cc|cc}
\toprule
\multirow{2}{*}{Method} & \multicolumn{2}{c|}{Val} & \multicolumn{2}{c}{Test} \\
 & gIoU & cIoU & gIoU & cIoU \\
\midrule
GRPO & 73.91\,{\scriptsize$\pm$0.65} & 73.71\,{\scriptsize$\pm$0.37} & 74.26\,{\scriptsize$\pm$0.67} & 76.19\,{\scriptsize$\pm$0.39} \\
\midrule
GRPO$\rightarrow$SFT & \textbf{76.21}\,{\scriptsize$\pm$0.06} & 76.75\,{\scriptsize$\pm$0.47} & 76.46\,{\scriptsize$\pm$0.07} & 78.46\,{\scriptsize$\pm$0.35} \\
SFT$\rightarrow$GRPO & 75.95\,{\scriptsize$\pm$0.03} & 76.35\,{\scriptsize$\pm$0.40} & 76.51\,{\scriptsize$\pm$0.11} & 78.75\,{\scriptsize$\pm$0.69} \\
BTO$\rightarrow$GRPO & 75.40\,{\scriptsize$\pm$0.44} & 75.38\,{\scriptsize$\pm$1.54} & 75.76\,{\scriptsize$\pm$0.04} & 77.05\,{\scriptsize$\pm$2.88} \\
\midrule
JPTO & 75.87\,{\scriptsize$\pm$0.04} & 74.52\,{\scriptsize$\pm$0.11} & \textbf{76.65}\,{\scriptsize$\pm$0.01} & 75.48\,{\scriptsize$\pm$0.80} \\
SFT$\rightarrow$JPTO & 75.89\,{\scriptsize$\pm$0.42} & 75.66\,{\scriptsize$\pm$2.30} & 76.15\,{\scriptsize$\pm$0.18} & 77.03\,{\scriptsize$\pm$1.29} \\
JPTO$\rightarrow$SFT & 75.96\,{\scriptsize$\pm$0.69} & 74.82\,{\scriptsize$\pm$1.30} & 76.61\,{\scriptsize$\pm$0.85} & 75.41\,{\scriptsize$\pm$2.16} \\
\midrule
GRTO \textit{(ours)} & 76.01\,{\scriptsize$\pm$0.34} & \textbf{77.25}\,{\scriptsize$\pm$0.38} & 76.49\,{\scriptsize$\pm$0.35} & \textbf{78.95}\,{\scriptsize$\pm$0.71} \\
\bottomrule
\end{tabular}
}%
\sbox{\tabboxB}{%
\begin{tabular}{l|cc|cc|cc|cc}
\toprule
\multirow{2}{*}{Method} & \multicolumn{2}{c|}{RS-X val$^\dagger$} & \multicolumn{2}{c|}{RS-X test} & \multicolumn{2}{c|}{RS-R*} & \multicolumn{2}{c}{RS*} \\
 & gIoU & cIoU & gIoU & cIoU & gIoU & cIoU & gIoU & cIoU \\
\midrule
GRPO & 63.3\,{\scriptsize$\pm$1.0} & 61.1\,{\scriptsize$\pm$2.6} & 59.9\,{\scriptsize$\pm$0.6} & 52.9\,{\scriptsize$\pm$1.0} & 74.9\,{\scriptsize$\pm$0.1} & 69.5\,{\scriptsize$\pm$0.2} & 72.3\,{\scriptsize$\pm$0.2} & 66.9\,{\scriptsize$\pm$1.2} \\
\midrule
GRPO$\rightarrow$SFT & 61.9\,{\scriptsize$\pm$0.4} & 53.6\,{\scriptsize$\pm$2.8} & 59.5\,{\scriptsize$\pm$0.1} & 52.4\,{\scriptsize$\pm$1.9} & 73.6\,{\scriptsize$\pm$0.8} & 67.0\,{\scriptsize$\pm$0.4} & 70.5\,{\scriptsize$\pm$1.0} & 61.3\,{\scriptsize$\pm$1.3} \\
SFT$\rightarrow$GRPO & 61.5\,{\scriptsize$\pm$0.4} & 53.5\,{\scriptsize$\pm$3.3} & 59.2\,{\scriptsize$\pm$0.2} & 53.2\,{\scriptsize$\pm$0.4} & 74.6\,{\scriptsize$\pm$0.0} & 69.3\,{\scriptsize$\pm$0.1} & 71.8\,{\scriptsize$\pm$0.1} & 63.8\,{\scriptsize$\pm$0.6} \\
BTO$\rightarrow$GRPO & 63.3\,{\scriptsize$\pm$0.4} & 60.6\,{\scriptsize$\pm$1.9} & 59.8\,{\scriptsize$\pm$0.6} & 53.2\,{\scriptsize$\pm$0.8} & \textbf{75.5}\,{\scriptsize$\pm$0.1} & 70.2\,{\scriptsize$\pm$0.2} & 72.6\,{\scriptsize$\pm$0.6} & 67.0\,{\scriptsize$\pm$0.0} \\
\midrule
JPTO & \textbf{64.0}\,{\scriptsize$\pm$0.1} & \textbf{62.0}\,{\scriptsize$\pm$0.2} & 59.5\,{\scriptsize$\pm$0.1} & 53.0\,{\scriptsize$\pm$0.7} & 75.3\,{\scriptsize$\pm$0.3} & 70.2\,{\scriptsize$\pm$0.0} & 72.5\,{\scriptsize$\pm$0.2} & 66.1\,{\scriptsize$\pm$0.4} \\
SFT$\rightarrow$JPTO & 61.6\,{\scriptsize$\pm$0.7} & 56.9\,{\scriptsize$\pm$0.9} & 59.7\,{\scriptsize$\pm$0.2} & 53.3\,{\scriptsize$\pm$0.7} & 74.5\,{\scriptsize$\pm$0.1} & 68.1\,{\scriptsize$\pm$0.9} & 71.2\,{\scriptsize$\pm$0.3} & 63.7\,{\scriptsize$\pm$0.8} \\
JPTO$\rightarrow$SFT & 62.4\,{\scriptsize$\pm$0.5} & 56.1\,{\scriptsize$\pm$0.4} & 59.9\,{\scriptsize$\pm$0.2} & 53.5\,{\scriptsize$\pm$0.2} & 74.9\,{\scriptsize$\pm$0.5} & 69.5\,{\scriptsize$\pm$1.3} & 72.0\,{\scriptsize$\pm$0.6} & 65.4\,{\scriptsize$\pm$0.2} \\
\midrule
GRTO \textit{(ours)} & 62.0\,{\scriptsize$\pm$0.5} & 56.9\,{\scriptsize$\pm$1.7} & \textbf{60.1}\,{\scriptsize$\pm$0.8} & \textbf{53.9}\,{\scriptsize$\pm$1.0} & 75.3\,{\scriptsize$\pm$0.1} & \textbf{70.5}\,{\scriptsize$\pm$0.8} & \textbf{72.7}\,{\scriptsize$\pm$0.4} & \textbf{67.6}\,{\scriptsize$\pm$0.5} \\
\bottomrule
\end{tabular}
}%
\pgfmathsetmacro{\tabRa}{\the\wd\tabboxA/(\the\ht\tabboxA+\the\dp\tabboxA)}%
\pgfmathsetmacro{\tabRb}{\the\wd\tabboxB/(\the\ht\tabboxB+\the\dp\tabboxB)}%
\pgfmathsetmacro{\tabH}{(\the\textwidth-3pt)/(\tabRa+\tabRb)}%
\pgfmathsetmacro{\tabSprev}{\tabH/(\the\ht\tabboxA+\the\dp\tabboxA)}%
\begin{subtable}[t]{\tabRa\dimexpr\tabH pt\relax}%
\centering
\caption{Remote sensing}
\label{tab:abl_er}
\resizebox*{!}{\tabH pt}{\usebox{\tabboxA}}%
\end{subtable}\hspace{2pt}%
\begin{subtable}[t]{\tabRb\dimexpr\tabH pt\relax}%
\centering
\caption{Reasoning segmentation}
\label{tab:abl_rs}
\resizebox*{!}{\tabH pt}{\usebox{\tabboxB}}%
\end{subtable}%
\end{table*}

\section{Full ablation results}\label{sec:ablations_per_set}

Table~\ref{tab:ablations} reports all ablations averaged over each domain's test sets, with every metric.

\begin{table}[t]
    \caption{Ablations across the three domains (the data of Figure~4 in the main paper). Values are two-seed means averaged over each domain's test sets (ReasonSeg: ReasonSeg, ReasonSeg-R, ReasonSeg-X; EarthReason: val, test; camouflage: CAMO, COD10K, NC4K). Best result per column in \textbf{bold}.}
    \label{tab:ablations}
    \centering
    \setlength{\tabcolsep}{2.5pt}
    \resizebox{\linewidth}{!}{
    \begin{tabular}{l|cc|cc|cccc}
         \toprule
         \multirow{2}{*}{Method} & \multicolumn{2}{c|}{ReasonSeg} & \multicolumn{2}{c|}{EarthReason} & \multicolumn{4}{c}{Camouflage} \\
         & gIoU & cIoU & gIoU & cIoU & S & wFm & E & MAE \\
         \midrule
         GRPO & 69.06 & 63.09 & 74.09 & 74.95 & .883 & .853 & .934 & .032 \\
         \midrule
         \multicolumn{9}{l}{\textit{Is joint optimization necessary?}} \\
         GRPO$\rightarrow$SFT & 67.88 & 60.22 & \textbf{76.33} & 77.60 & .890 & .868 & .939 & .029 \\
         SFT$\rightarrow$GRPO & 68.52 & 62.08 & 76.23 & 77.55 & .894 & .871 & .943 & .029 \\
         BTO$\rightarrow$GRPO & 69.29 & 63.48 & 75.58 & 76.21 & .895 & .872 & .944 & .028 \\
         \midrule
         \multicolumn{9}{l}{\textit{Is tool bootstrapping necessary?}} \\
         JPTO & 69.09 & 63.08 & 76.26 & 75.00 & .895 & .874 & .945 & .028 \\
         SFT$\rightarrow$JPTO & 68.45 & 61.71 & 76.02 & 76.35 & \textbf{.898} & \textbf{.877} & .946 & \textbf{.027} \\
         JPTO$\rightarrow$SFT & 68.95 & 62.78 & 76.28 & 75.11 & .896 & .875 & .946 & .028 \\
         \midrule
         GRTO & \textbf{69.38} & \textbf{64.02} & 76.25 & \textbf{78.10} & \textbf{.898} & \textbf{.877} & \textbf{.947} & \textbf{.027} \\
         \bottomrule
    \end{tabular}
    }
\end{table}

Table~\ref{tab:abl} reports all configurations of the ablation study on every evaluated split, including the ReasonSeg-X validation split used for checkpoint selection.

\section{Detailed experimental setup}\label{sec:detailed_experimental_setup}

\noindent\textbf{Configurations.} All configurations start from the same policy $\pi_{\theta_0}$ and tool $\psi_{\omega_0}$ and differ only in what is trained and in which order:
\begin{itemize}
    \item \textbf{GRPO}: the policy is trained with GRPO; the tool is frozen.
    \item \textbf{BTO}: the tool is trained on a replay buffer of rollouts generated by the frozen $\pi_{\theta_0}$; each prompt's group is generated once and reused in later epochs. The policy is frozen.
    \item \textbf{JPTO}: policy and tool are trained jointly with Eq.~(6) in the main paper.
    \item \textbf{GRTO}: JPTO initialized from a BTO checkpoint.
    \item \textbf{SFT}: as BTO, with the same replay buffer, but the tool loss is the plain mean over the valid rollouts of each group, without the BTO weights.
\end{itemize}
The ablations without joint optimization combine GRPO with a separately trained tool:
\begin{itemize}
    \item \textbf{GRPO$\rightarrow$SFT}: the policy is frozen at the selected GRPO checkpoint, and the tool is fine-tuned with SFT on a replay buffer of this policy's rollouts, of the same size as the BTO buffer.
    \item \textbf{SFT$\rightarrow$GRPO}: GRPO with the tool frozen at an SFT checkpoint.
    \item \textbf{BTO$\rightarrow$GRPO}: GRPO with the tool frozen at the same BTO checkpoint as GRTO.
\end{itemize}
The ablations without bootstrapping keep JPTO and change how the tool is trained beforehand or afterwards:
\begin{itemize}
    \item \textbf{JPTO}: as above, from the frozen tool.
    \item \textbf{SFT$\rightarrow$JPTO}: JPTO initialized from an SFT checkpoint, at the same epoch as the BTO checkpoint of GRTO.
    \item \textbf{JPTO$\rightarrow$SFT}: as GRPO$\rightarrow$SFT, but from the selected JPTO checkpoint.
\end{itemize}

\noindent\textbf{Common settings.} We use AdamW ($\beta_1{=}0.9$, $\beta_2{=}0.999$, $\epsilon{=}10^{-8}$, no weight decay) with a constant learning rate and no warmup, bf16, and DeepSpeed ZeRO-2. Rollouts are sampled with temperature 0.9 and top-$k$ 50. Advantages are normalized as $(R_i-\mu)/\sigma$, the PPO clipping range is $\varepsilon{=}0.2$, and summands of Eq.~(3) in the main paper are divided by the maximum output length instead of each rollout's length to avoid length bias~\cite{liu2025understanding}. Each sampled group is used for one optimizer step. The KL weight is $\beta{=}0.01$ for all policy-training stages. We save a checkpoint after every epoch and report the checkpoint selected on the validation split. Per-domain settings are listed in Table~\ref{tab:setup}.

\begin{table*}[t]
\centering
\caption{Per-domain training settings. LR: learning rate. SFT: every SFT stage (SFT$\rightarrow$GRPO, SFT$\rightarrow$JPTO, GRPO$\rightarrow$SFT, JPTO$\rightarrow$SFT). JPTO$^2$: the JPTO stage of GRTO (and of SFT$\rightarrow$JPTO).}
\label{tab:setup}
\resizebox{\linewidth}{!}{
\begin{tabular}{llll}
\toprule
 & Reasoning segmentation & Remote sensing & Camouflage detection \\
\midrule
Policy & Qwen3-VL-8B, StAR stage-1 init~\cite{yun2026star} & InternVL3.5-8B & InternVL3.5-8B \\
Tool & SAM2.1-hiera-large & SAM3 & SAM3 \\
Trained policy weights & LoRA (rank 64, $\alpha$ 64, all linear layers) & full model & full model \\
Trained tool weights & mask decoder & mask decoder & mask decoder \\
Policy LR & $5\times10^{-6}$ & $1\times10^{-6}$ & $1\times10^{-6}$ \\
Tool LR & BTO, SFT: $1\times10^{-4}$; JPTO, JPTO$^2$: $1\times10^{-6}$ & JPTO, BTO, SFT: $1\times10^{-4}$; JPTO$^2$: $1\times10^{-6}$ & $1\times10^{-4}$ \\
Prompts $\times$ group size per step & $16\times64$ (REST: top-8 + bottom-8 for the policy, top-8 valid for the tool) & $64\times8$ & $64\times8$ \\
Replay-buffer group size & 128 & 8 & 8 \\
Max.\ output tokens & 2048 & 1024 & 1024 \\
Gradient clipping & 1.0 & 1.0 & 1.0 \\
Training budget & 10 epochs (15 steps each) & 16 epochs (38 steps each) & 16 epochs (61 steps each) \\
Checkpoint selection & best mean(gIoU, cIoU) on ReasonSeg-X val & best mean(gIoU, cIoU) on 200 val questions & best wFm on 135 val images \\
\bottomrule
\end{tabular}
}
\end{table*}

\noindent\textbf{Reasoning segmentation.} We post-train StAR's stage-1 checkpoint~\cite{yun2026star} with REST: of the 64 rollouts per prompt, the policy is updated on the 8 highest- and 8 lowest-advantage ones, and in JPTO the tool on the 8 highest-reward valid ones. We train on the 240 ReasonSeg-X training samples, select checkpoints within the 10-epoch budget on the 156 validation samples using greedy decoding, and test on ReasonSeg-X (773), ReasonSeg-R (968), and ReasonSeg (779) test samples. Images are resized to $840\times840$ for the policy. At test time, we use majority voting over 32 samples (temperature 1.0, top-$p$ 0.9), following StAR \cite{yun2026star}.

\noindent\textbf{Remote sensing.} EarthReason has 2,371 training, 1,135 validation, and 1,928 test images with ${\sim}6$ questions each (14,225, 6,802, and 11,568 questions). Each epoch samples one question per training image; the replay buffer therefore contains one group per sampled image--question pair. We select checkpoints on 200 validation questions (one per image) and test on the full validation and test sets.

\noindent\textbf{Camouflage detection.} We hold out 100 COD10K and 35 CAMO training images for validation and train on the remaining 3,904 images of both training sets. We test on COD10K (2,026), CAMO (250), and zero-shot on NC4K (4,121). The prompt is static (Section~\ref{sec:prompts_used}), so each image has a single replay-buffer group.

\noindent\textbf{Reward.} The GRPO reward is
\begin{equation}
R = 0.9\,R_{\mathrm{IoU}} + 0.1\,R_{\mathrm{format}},
\end{equation}
where $R_{\mathrm{IoU}}$ is the IoU between the union of the predicted masks and the union of the ground-truth masks,
\begin{equation}
R_{\mathrm{IoU}} = \frac{\left|\hat{M}\cap M\right|}{\left|\hat{M}\cup M\right|},
\end{equation}
and $R_{\mathrm{format}}=1$ if the policy outputs a parsable prompt for the tool or an explicit empty answer, and 0 otherwise.

\noindent\textbf{Tool loss.} For each prompt, the tool predicts mask logits $\hat M$ with probabilities $S=\sigma(\hat M)\in[0,1]^{H\times W}$. With the binary ground-truth mask $M\in\{0,1\}^{H\times W}$, the loss is $\mathcal{L}=\mathcal{L}_{\mathrm{BCE}}+\mathcal{L}_{\mathrm{sIoU}}$ with
\begin{equation}
\begin{split}
\mathcal L_{\mathrm{BCE}}
=
-\frac{1}{HW}
\sum_{u=1}^{H}\sum_{v=1}^{W}
\big[
&M_{uv}\log S_{uv}
\\
&+
(1-M_{uv})\log(1-S_{uv})
\big],
\end{split}
\end{equation}
\begin{equation}
\mathcal L_{\mathrm{sIoU}}
=
1-
\frac{
\sum_{u,v} S_{uv}M_{uv}
}{
\sum_{u,v}
\left(S_{uv}+M_{uv}-S_{uv}M_{uv}\right)
}.
\end{equation}

\noindent\textbf{Hyperparameter search.} Before the final runs, we varied one setting at a time from a default configuration, trained for a few epochs, and monitored validation trends (Table~\ref{tab:hyperparameter_search}). Settings established by a comparable baseline, such as StAR's for reasoning segmentation, were kept fixed.

\begin{table}[t]
\centering
\caption{Hyperparameter settings explored during the initial search.}
\label{tab:hyperparameter_search}
\begin{tabular}{@{}l>{\raggedright\arraybackslash}p{0.46\linewidth}@{}}
\toprule
\textbf{Hyperparameter} & \textbf{Values explored} \\
\midrule
Advantage normalization & no ($R_i$), only mean ($R_i-\mu$), yes ($\frac{R_i-\mu}{\sigma}$) \\
Objective normalization & rollout length $\frac{1}{|o_i|}$, constant $\frac{1}{L}$ \\
SAM learning rate & $1 \times 10^{-6}$, $1 \times 10^{-5}$, $5 \times 10^{-5}$, $1 \times 10^{-4}$ \\
$\beta$ KL regularization & 0, 0.001, 0.01, 1 \\
SAM training mode & decoder only, decoder + LoRA encoder, full fine-tuning \\
LoRA rank & 64, 128, 256, no LoRA \\
Gradient clipping & with, without \\
Exponential moving average & with, without \\
\bottomrule
\end{tabular}
\end{table}

\section{Statistical significance}\label{sec:significance}

We test whether GRTO's improvement over GRPO exceeds the uncertainty due to the finite evaluation sets. For each metric, we compute the difference between GRTO and GRPO with a paired bootstrap over evaluation images ($B{=}10{,}000$): within each test set, images are resampled with replacement (all queries about one image together, as EarthReason has ${\sim}6$ queries per image), the same resample is applied to both methods and seeds, and every metric is recomputed exactly as in the official evaluation (\eg, cIoU from cumulative intersections and unions). Differences are averaged over the two seeds and over each domain's test sets. The intervals are conditional on the trained checkpoints, \ie, they quantify evaluation-set uncertainty, not variability over training runs. We test one hypothesis per domain, GRTO vs.\ GRPO on mean(gIoU, cIoU) for ReasonSeg and EarthReason and on wFm, the checkpoint-selection metric, for camouflage, with Holm correction~\cite{holm1979simple}.

\begin{table}[t]
\centering
\caption{GRTO $-$ GRPO with 95\% paired bootstrap confidence intervals over evaluation images, mean over two seeds and each domain's test sets (in brackets: each seed's difference). Camouflage measures $\times 100$; MAE as reduction (positive is better). $p$: Holm-adjusted bootstrap $p$-value of the primary comparison (marked $^\star$).}
\label{tab:significance}
\setlength{\tabcolsep}{3pt}
\resizebox{\linewidth}{!}{
\begin{tabular}{llccc}
\toprule
Domain & Metric & GRTO $-$ GRPO [95\% CI] & per seed & $p$ \\
\midrule
\multirow{3}{*}{ReasonSeg} & mean(gIoU, cIoU)$^\star$ & $+0.63$ [$-0.00$, $+1.24$] & $+1.06$ / $+0.19$ & 0.051 \\
 & gIoU & $+0.32$ [$-0.05$, $+0.68$] & $+0.54$ / $+0.10$ & \\
 & cIoU & $+0.94$ [$-0.10$, $+1.97$] & $+1.58$ / $+0.29$ & \\
\midrule
\multirow{3}{*}{EarthReason} & mean(gIoU, cIoU)$^\star$ & $+2.63$ [$+2.03$, $+3.25$] & $+2.18$ / $+3.09$ & $<$0.001 \\
 & gIoU & $+2.16$ [$+1.77$, $+2.55$] & $+1.44$ / $+2.87$ & \\
 & cIoU & $+3.11$ [$+2.14$, $+4.13$] & $+2.91$ / $+3.30$ & \\
\midrule
\multirow{6}{*}{Camouflage} & wFm$^\star$ & $+2.31$ [$+1.81$, $+2.82$] & $+2.18$ / $+2.45$ & $<$0.001 \\
 & S & $+1.52$ [$+1.17$, $+1.88$] & $+1.47$ / $+1.56$ & \\
 & E & $+1.34$ [$+0.90$, $+1.80$] & $+1.30$ / $+1.39$ & \\
 & MAE (reduction) & $+0.51$ [$+0.32$, $+0.72$] & $+0.47$ / $+0.55$ & \\
 & gIoU & $+2.63$ [$+2.11$, $+3.17$] & $+2.54$ / $+2.72$ & \\
 & cIoU & $+3.56$ [$+2.50$, $+4.55$] & $+3.16$ / $+3.97$ & \\
\bottomrule
\end{tabular}
}
\end{table}

Table~\ref{tab:significance} reports the results. On EarthReason and camouflage, GRTO's improvement is significant for every metric ($p<0.001$ for the primary comparisons) and positive in each seed and on every test set. On ReasonSeg, the improvement is smaller and marginally significant ($p=0.051$), positive in both seeds and on all three test sets. This is consistent with the analysis in Section~4.2 of the main paper: natural images exhibit the smallest domain shift of the frozen SAM (Figure~5a in the main paper), and fine-tuning SAM on pseudo-GT prompts yields the smallest gain there (Section~\ref{sec:gt_prompts}), leaving little headroom for tool adaptation. In addition, ReasonSeg-X provides only 240 training queries and we start from a state-of-the-art checkpoint evaluated with majority voting, so both the attainable gain and its estimate are more sensitive to training randomness.

\section{Additional measures for decoder domain shift}\label{sec:gt_prompts}

We verify that the relation between the tool's domain shift and GRTO's gains (Figure~5a in the main paper) does not depend on the choice of model or measure, using two further measures.

\noindent\textbf{GBC on SAM2.1.} We repeat the GBC analysis with the prompt-free image embedding of SAM2.1 on all test images. GBC values are not comparable across models, since they are computed in different feature spaces, but the domains can be compared within each model.

\noindent\textbf{Decoder improvement under pseudo-ground-truth prompts.} To measure the gap in mask quality rather than in features, we remove the reasoning from the task. In line with common practice, we evaluate SAM with prompts derived from the ground-truth annotations, which we call pseudo-ground-truth (pseudo-GT) prompts~\cite{kirillov2023sam,ren2024segment}: one bounding box around the whole ground-truth target, plus one positive point at its most interior pixel for ReasonSeg (SAM2.1) and the class name for EarthReason and COD10K (SAM3; CAMO and NC4K provide no class names, so the box is used alone). We then compare the pretrained SAM with a SAM adapted to the target domain under identical prompts~\cite{zhang2024improving,archit2025segment}: we fine-tune SAM with SFT on pseudo-GT prompts of the single-region training targets and compare it to the frozen SAM on the single-region test targets (Section~4.1 of the main paper). The gain of the fine-tuned SAM measures the error that the frozen decoder makes even when the target is fully specified.

\begin{table}[t]
\centering
\caption{Measures of the domain shift per domain (mean over its test sets), next to GRTO's relative gain over GRPO. GBC values are comparable across domains within a model, not across models. SFT gain: relative gain (\%) of SAM fine-tuned on pseudo-GT prompts over the frozen SAM, mean of the gIoU and cIoU gains, single-region targets.}
\label{tab:gt_prompts}
\setlength{\tabcolsep}{3pt}
\resizebox{\linewidth}{!}{
\begin{tabular}{lccc}
\toprule
 & ReasonSeg & EarthReason & Camouflage \\
\midrule
SAM3 domain shift ($-$GBC) & 0.055 & 0.067 & 0.092 \\
SAM2.1 domain shift ($-$GBC) & 0.026 & 0.044 & 0.045 \\
SFT SAM gain, pseudo-GT prompts (\%) & 1.7 & 4.9 & 3.1 \\
\midrule
GRTO $-$ GRPO (\%) & 1.0 & 3.6 & 3.9 \\
\bottomrule
\end{tabular}
}
\end{table}

\begin{figure}[t]
    \centering
    \includegraphics[width=\linewidth]{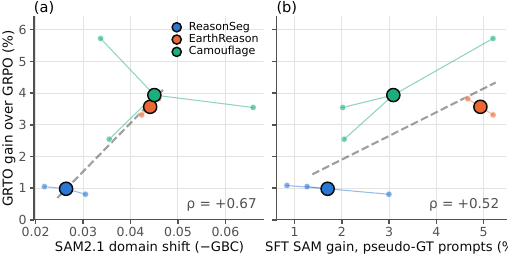}
    \caption{GRTO's relative gain over GRPO against (a) the SAM2.1 domain shift and (b) the gain of SAM fine-tuned on pseudo-GT prompts over the frozen SAM. Small markers: test sets, joined to their domain mean (large markers); dashed: fit through the domain means; $\rho$: Spearman correlation over the eight test sets.}
    \label{fig:gt_prompts}
\end{figure}

Table~\ref{tab:gt_prompts} and Figure~\ref{fig:gt_prompts} show the results. With SAM2.1, the domains follow the same order as with SAM3, ReasonSeg having the smallest and camouflage the largest domain shift, and GRTO's gain correlates with the gap across test sets ($\rho=0.67$, compared with $0.62$ for SAM3). Fine-tuning on pseudo-GT prompts also improves the frozen SAM least on ReasonSeg, where GRTO's gains are smallest, and its gain correlates with GRTO's gain across test sets ($\rho=0.52$).

\section{Policy divergence and replay-buffer analysis}\label{sec:bto_analysis}

KL-regularized RL, including BTO, is solvable without sophisticated exploration if groups contain positive rewards, $\mathbb{E}[r]\geq v>0$; even sampling from the reference policy suffices~\cite{brantley2025acceleratingrlllmreasoning,yue2025does}. BTO therefore introduces no assumptions beyond those already required for standard RL post-training. The same estimator of the optimal value is used by A*PO~\cite{brantley2025acceleratingrlllmreasoning} and OAPL~\cite{ritter2026llmslearnreasonoffpolicy} in a different setting, and A*PO provides performance guarantees for training under it~\cite{brantley2025acceleratingrlllmreasoning}.

\noindent\textbf{Replay-buffer quality.} We verify this condition on our replay buffers (Table~\ref{tab:buffer_quality}). A rollout is valid if it parses into a usable spatial prompt. Nearly all groups contain valid and diverse rollouts, and BTO further reduces the share of groups without a useful rollout.

\begin{table}[t]
\centering
\caption{Replay-buffer quality. $G$ rollouts per prompt from $\pi_{\theta_0}$ ($G{=}8$ for EarthReason and camouflage, $G{=}64$ for ReasonSeg-X). Rewards are recomputed after BTO; validity and diversity are properties of the fixed buffer.}
\label{tab:buffer_quality}
\setlength{\tabcolsep}{3pt}
\resizebox{\linewidth}{!}{
\begin{tabular}{lccc}
\toprule
Metric & EarthReason & Camouflage & ReasonSeg-X \\
\midrule
Groups with no valid rollout & 1.00\% & 0.20\% & 0.00\% \\
Groups with $\max r < 0.11$, before BTO & 8.00\% & 1.60\% & 3.50\% \\
Groups with $\max r < 0.11$, after BTO  & 6.80\% & 1.00\% & 3.50\% \\
Groups with zero reward variance, before BTO & 1.40\% & 0.60\% & 3.00\% \\
Groups with zero reward variance, after BTO  & 1.20\% & 0.80\% & 3.00\% \\
Mean distinct strings per $G$ rollouts & 7.65 / 8 & 8.00 / 8 & 63.99 / 64 \\
Exact duplicate rate & 4.32\% & 0.00\% & 0.01\% \\
\bottomrule
\end{tabular}
}
\end{table}

\noindent\textbf{Policy divergence.} BTO trains the tool for the posterior policy $\pi^*=\pi_{\theta^*(\omega)}$ estimated from the reference policy. Using the same evaluation prompts and KL estimator, we compare the trained policies with $\pi_{\theta_0}$ and with $\pi^*$ (Table~\ref{tab:policy_kl}). For every trained policy, the divergence from $\pi^*$ is much smaller than from $\pi_{\theta_0}$: the estimated posterior is substantially closer to the learned policy than the reference policy is. Moreover, BTO decreases $\text{D}_\text{KL}(\pi_{\theta_0}\|\pi^*)$ in all domains, moving the posterior closer to the policy's existing output distribution. An exact match is unnecessary, since BTO only initializes the tool and the subsequent JPTO stage adapts it to fresh rollouts.

\begin{table}[t]
\centering
\caption{Sequence-level KL divergence between trained policies $\pi_{\theta}$ and reference $\pi_{\theta_0}$ or the estimated posterior $\pi_{\theta^*}$ ($\beta{=}0.01$), on the same evaluation prompts. Last row: $\text{D}_\text{KL}(\pi_{\theta_0}\|\pi_{\theta^*})$ before $\rightarrow$ after BTO.}
\label{tab:policy_kl}
\setlength{\tabcolsep}{3pt}
\resizebox{\linewidth}{!}{
\begin{tabular}{lcccccc}
\toprule
& \multicolumn{2}{c}{EarthReason} & \multicolumn{2}{c}{Camouflage} & \multicolumn{2}{c}{ReasonSeg-X} \\
\cmidrule(lr){2-3}\cmidrule(lr){4-5}\cmidrule(lr){6-7}
Policy & $\pi_{\theta}\|\pi_{\theta_0}$ & $\pi_{\theta}\|\pi_{\theta^*}$ & $\pi_{\theta}\|\pi_{\theta_0}$ & $\pi_{\theta}\|\pi_{\theta^*}$ & $\pi_{\theta}\|\pi_{\theta_0}$ & $\pi_{\theta}\|\pi_{\theta^*}$ \\
\midrule
GRPO   & \multicolumn{2}{c}{14.43 $\,>\,$ 4.90\,} &  \multicolumn{2}{c}{16.74 $\,>\,$ 15.03\,}& \multicolumn{2}{c}{42.16 $\,>\,$ 19.26\,} \\
BTO$\rightarrow$GRPO & \multicolumn{2}{c}{13.93 $\,>\,$ 0.90\,} & \multicolumn{2}{c}{10.27 $\,>\,$ 7.42\,\,\,} & \multicolumn{2}{c}{39.23 $\,>\,$ 18.75\,}\\
GRTO & \multicolumn{2}{c}{11.31 $\,>\,$ 1.03\,} & \multicolumn{2}{c}{13.23 $\,>\,$ 11.30\,} & \multicolumn{2}{c}{42.71 $\,>\,$ 18.40\,} \\
\midrule
$\pi_{\theta_0}\|\pi^*$ & \multicolumn{2}{c}{14.13 $\rightarrow$ 13.26} & \multicolumn{2}{c}{9.50 $\rightarrow$ 7.54} & \multicolumn{2}{c}{17.79 $\rightarrow$ 17.64} \\
\bottomrule
\end{tabular}
}
\end{table}

\section{Effect of bootstrapping on convergence}\label{sec:convergence}

\begin{figure}[t]
    \centering
    \includegraphics[width=\linewidth]{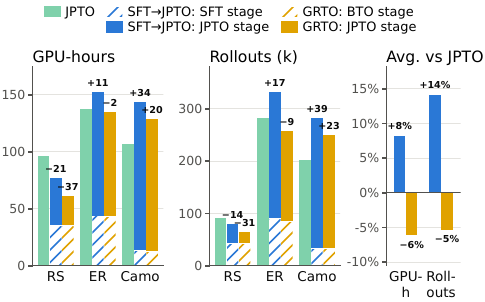}
    \caption{Training cost until the selected checkpoint for JPTO, SFT$\rightarrow$JPTO and GRTO, two-seed means. Labels: change relative to JPTO (\%). Avg.: mean change relative to JPTO over the three domains. RS: ReasonSeg; ER: EarthReason; Camo: camouflage.}
    \label{fig:convergence}
\end{figure}

Figure~\ref{fig:convergence} compares the training cost until the selected checkpoint of the three best-performing configurations: JPTO, SFT$\rightarrow$JPTO and GRTO, including the tool stage of the latter two. We measure GPU-hours and policy rollouts. The SFT stage replays the same buffer as the BTO stage, so we charge it the BTO stage's rollout cost. GRTO converges faster than JPTO in two of the three domains, ReasonSeg ($-37\%$ GPU-hours, $-31\%$ rollouts) and EarthReason ($-2\%$, $-9\%$), and is slower on camouflage ($+20\%$, $+23\%$), saving on average 6\% of GPU-hours and 5\% of rollouts. Replacing BTO with plain SFT on the same buffer removes this advantage: the tool stage costs about the same, but the subsequent JPTO stage needs more steps to reach its selected checkpoint. SFT$\rightarrow$JPTO is therefore slower than GRTO in every domain (ReasonSeg $+26\%$ GPU-hours, $+24\%$ rollouts; EarthReason $+13\%$, $+28\%$; camouflage $+11\%$, $+12\%$) and costs on average 8\% more GPU-hours and 14\% more rollouts than JPTO.

\section{Comparison to pure SFT}\label{sec:sft_comparison}

In camouflage detection and remote sensing, SAM cannot be trained with SFT alone, as the queries are too complex to be passed to SAM in their raw form. On simpler referring segmentation benchmarks, the RefCOCO family~\cite{yu2016refcoco,mao2016refcocog}, this is possible. For a targeted comparison against SFT baselines on the RefCOCO family, we refer to Table~S.4 of StAR~\cite{yun2026star}, which reports four pure SFT baselines trained extensively on the RefCOCO training sets, and evaluate our reasoning segmentation GRTO model on the same benchmarks (Table~\ref{tab:refcoco}). We do not use majority voting here, as the test sets contain too many samples for it to be computationally feasible.

\begin{table}[t]
\centering
\caption{Reported cIoU on the RefCOCO family. An asterisk (*) denotes methods trained extensively on the RefCOCO training sets. Best result per column in \textbf{bold}.}
\label{tab:refcoco}
\setlength{\tabcolsep}{4pt}
\resizebox{\linewidth}{!}{
\begin{tabular}{lccc}
\toprule
\multirow{2}{*}{Model} & RefCOCO & RefCOCO+ & RefCOCOg \\
 & testA & testA & test \\
\midrule
LISA-7B*~\cite{lai2023lisa} (SFT) & 76.5 & 67.4 & 68.5 \\
SegLLM*~\cite{xu2024segllm} (SFT) & \textbf{81.5} & 73.0 & 73.6 \\
READ*~\cite{qian2025reasoning} (SFT) & 80.2 & 73.7 & 71.4 \\
CoReS*~\cite{bao2024cores} (SFT) & 78.6 & 70.0 & 70.7 \\
StAR Qwen2.5~\cite{yun2026star} & 78.6 & 74.0 & 71.0 \\
StAR Qwen3~\cite{yun2026star} & 78.7 & 74.4 & 73.3 \\
\midrule
StAR Qwen3 + GRTO \textit{(ours)} & 79.9 & \textbf{74.5} & \textbf{73.9} \\
\bottomrule
\end{tabular}
}
\end{table}

Our model has seen only a small subset of the RefCOCO family during its first-stage pretraining, before the subsequent REST, BTO, and JPTO stages, whereas the SFT baselines are trained extensively on the complete RefCOCO training sets. Nevertheless, StAR Qwen3 + GRTO remains competitive, surpassing all SFT baselines on two of the three benchmarks. In this near-zero-shot setting, GRTO consistently improves the underlying StAR Qwen3 model, and achieves the highest score on RefCOCOg, which contains the most complex referring expressions (8.43 words on average, versus 3.61 for RefCOCO and 3.53 for RefCOCO+).

\section{Qualitative examples}\label{sec:qualitative_examples}

\begin{figure*}[t]
    \centering
    \includegraphics[width=\linewidth]{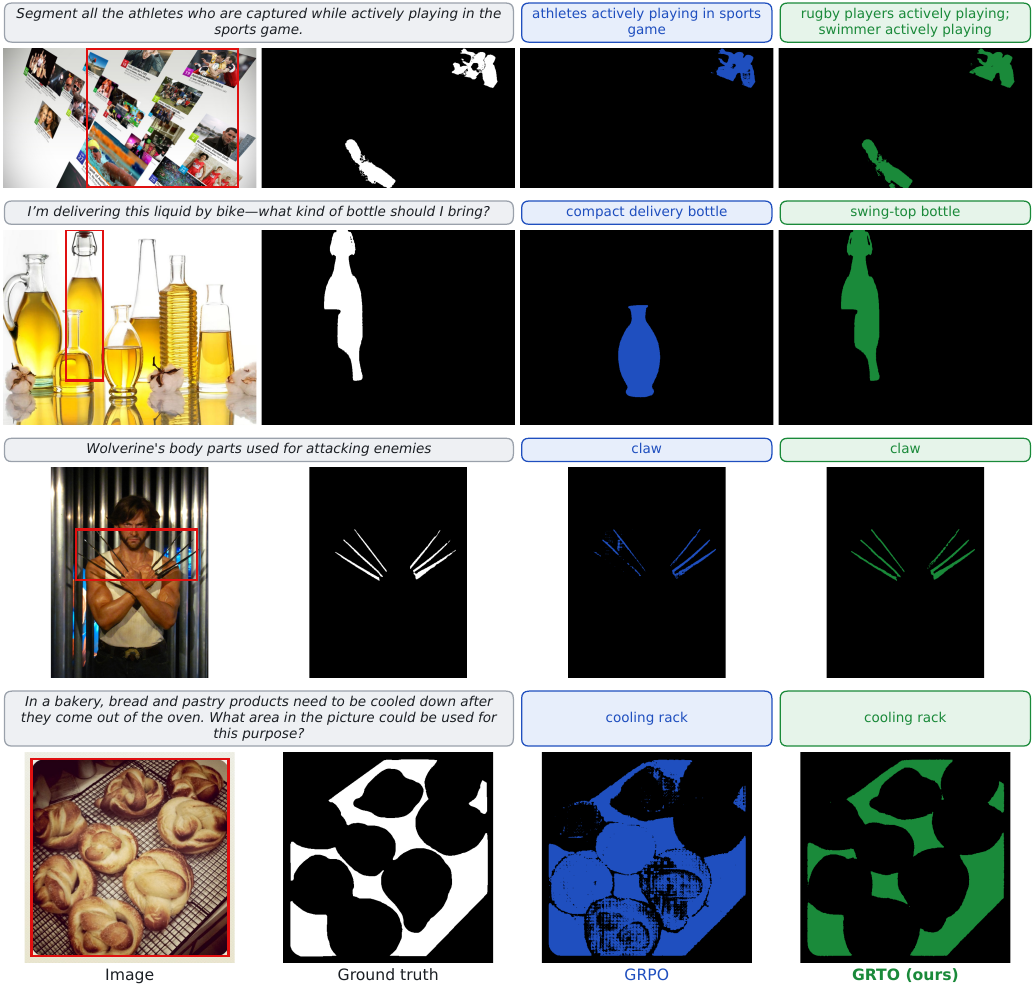}
    \caption{Reasoning segmentation test examples (rows 1--2: ReasonSeg-X; rows 3--4: ReasonSeg-R and ReasonSeg). Above the image: query; above the masks: prompt passed to the tool by GRPO (blue) and GRTO (green). The red box marks the ground-truth target.}
    \label{fig:qual_reasonseg}
\end{figure*}

\begin{figure*}[t]
    \centering
    \includegraphics[width=\linewidth]{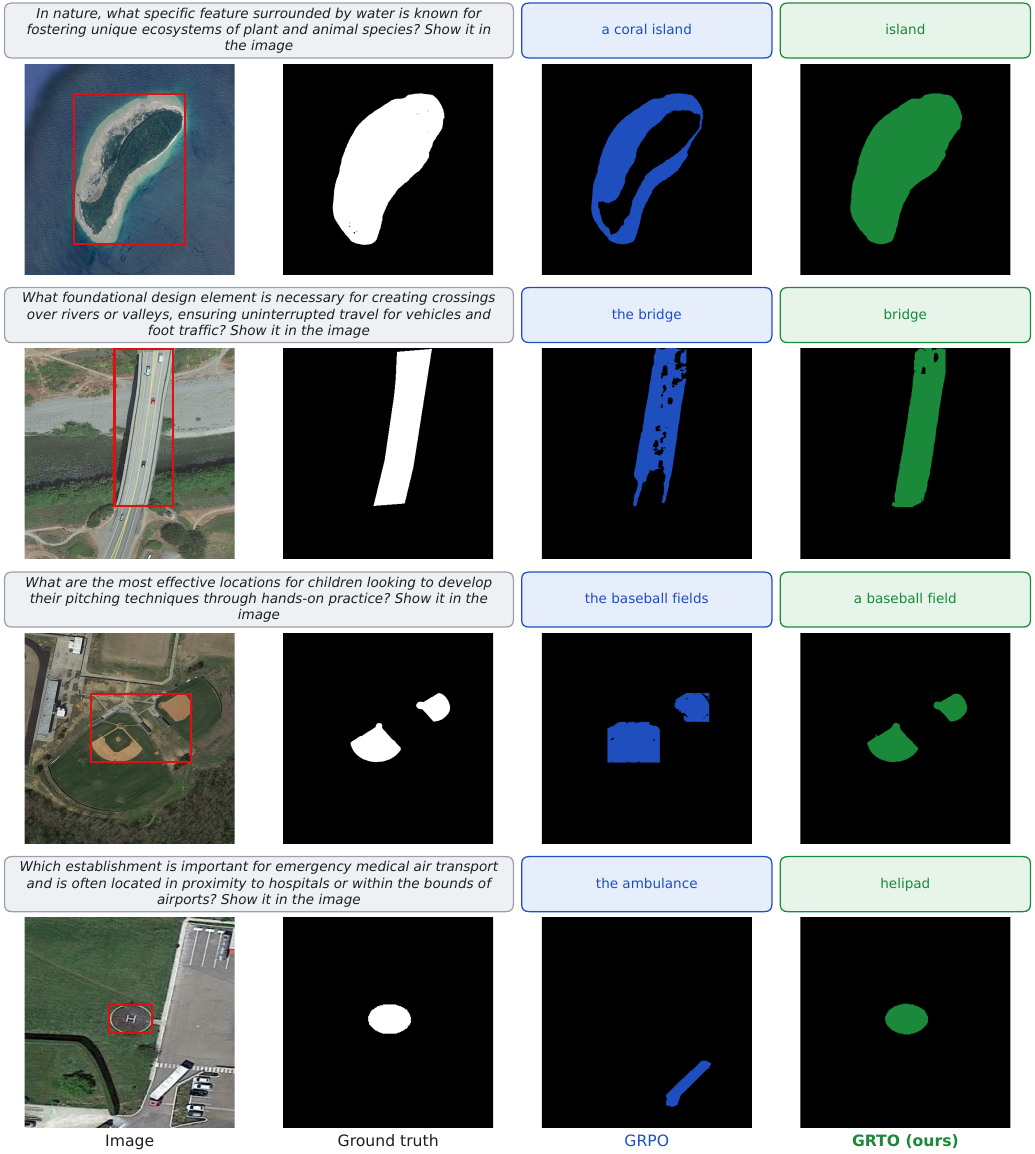}
    \caption{EarthReason test examples. Above the image: query; above the masks: prompt passed to the tool by GRPO (blue) and GRTO (green). The red box marks the ground-truth target.}
    \label{fig:qual_earthreason}
\end{figure*}

\begin{figure*}[t]
    \centering
    \includegraphics[width=\linewidth]{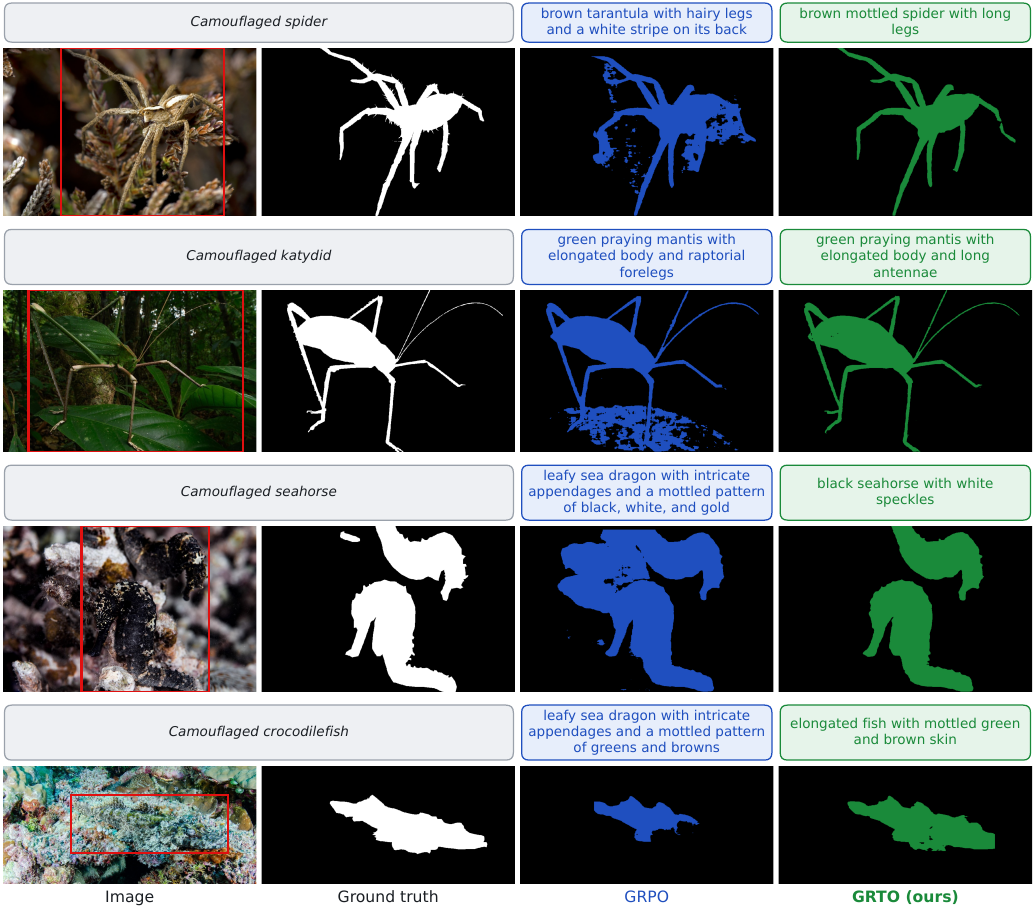}
    \caption{COD10K test examples. Above the image: ground-truth class; above the masks: prompt passed to the tool by GRPO (blue) and GRTO (green). The red box marks the ground-truth target.}
    \label{fig:qual_camouflage}
\end{figure*}

Figures~\ref{fig:qual_reasonseg}, \ref{fig:qual_earthreason}, and~\ref{fig:qual_camouflage} show test examples. For reasoning segmentation, we report the reasoning of both policies below; the shown output is the majority-voting sample whose box best matches the final mask.

\noindent\textbf{Figure~\ref{fig:qual_reasonseg}, row 1} (``Segment all the athletes who are captured while actively playing in the sports game.'').

\textcolor{grpoblue}{\textit{Looking at the image, I need to find athletes who are captured while actively playing in a sports game. Analyzing the various thumbnails: 1. In the top right thumbnail, there's a rugby match with players actively engaged in play - one player is holding the ball while being tackled by others. 2. Looking at other thumbnails, I can see swimming, movie reviews, group photos, etc., but only the rugby scene shows athletes actively playing in a sports game. The rugby players in the top right thumbnail are clearly captured mid-game action - one player is holding the ball while being tackled by opponents. This matches the query perfectly.} (GRPO; prompt: ``athletes actively playing in sports game'')}

\textcolor{grtogreen}{\textit{Looking at the image, I need to find athletes who are captured while actively playing in a sports game. Analyzing the various thumbnails: 1. There's a thumbnail showing a rugby match with players in action (Australia vs South Africa) 2. There's a thumbnail showing a swimmer in a pool, which is a sports activity 3. There's a thumbnail showing a crowd at what appears to be a sporting event The most obvious sports athletes actively playing would be: - The rugby players in the top right thumbnail (Australia vs South Africa) - The swimmer in the bottom center thumbnail (The Spirit of Australia) The rugby players are clearly engaged in a game, with one player holding the ball and being tackled. The swimmer is also actively participating in a sporting event.} (GRTO; prompt: ``rugby players actively playing; swimmer actively playing'')}

\noindent\textbf{Figure~\ref{fig:qual_reasonseg}, row 2} (``I'm delivering this liquid by bike---what kind of bottle should I bring?'').

\textcolor{grpoblue}{\textit{The query "i'm delivering this liquid by bike---what kind of bottle should i bring" is asking for a practical, portable bottle suitable for bicycle delivery. Looking at the image, we have several glass bottles of different shapes and sizes. For bike delivery, one would want a lightweight, durable, and easy-to-handle bottle. The most practical choice from the image would be a bottle that is not too tall or bulky, with a secure cap to prevent leaks, and ideally has a handle or is easy to grip. Among the bottles shown: - The large carafe on the left has a handle and is relatively wide at the base, which could be heavy and unwieldy for bike delivery. - The tall, narrow bottles (like the one in the center-right with ridges) are very tall and could be unstable or difficult to carry. - The smaller, more compact bottles with narrow necks (like the one in the center) might be more manageable. - The bottle with the swing-top closure (the one with the metal clasp) is secure and may be easier to handle. The most suitable bottle for bike delivery would likely be the smaller, more compact bottle with a secure cap and manageable size. The bottle in the center, which is smaller and has a relatively wide mouth but a secure-looking neck, seems ideal for this purpose.} (GRPO; prompt: ``compact delivery bottle'')}

\textcolor{grtogreen}{\textit{Looking at the query "i'm delivering this liquid by bike---what kind of bottle should i bring", I need to think about what type of bottle would be practical for bicycle delivery. For bike delivery, you'd want a lightweight, durable, and easy-to-handle bottle that won't break easily and can be secured to a bike. Among the bottles shown, the most practical for bike delivery would likely be the smaller, more compact bottles that are easier to carry and mount on a bike. The bottle with the swing-top lid (the tall one in the center-left) would be good as it's sealed and doesn't require a cap, but it's still fairly large. The smaller, more slender bottles (like the one in the center-right) might be more suitable for bike delivery as they're lighter and more compact. However, the most practical choice would be a bottle that is lightweight, has a secure closure, and is easy to handle while riding. The bottle with the swing-top lid (the tall one with the metal clasp) seems ideal because it's sealed when closed, won't leak, and can be securely fastened to a bike rack or handlebars.} (GRTO; prompt: ``swing-top bottle'')}

\noindent\textbf{Figure~\ref{fig:qual_reasonseg}, row 3} (``Wolverine's body parts used for attacking enemies'').

\textcolor{grpoblue}{\textit{Looking at the image, I need to identify "Wolverine's body parts used for attacking enemies". Based on the character's appearance, this clearly refers to his adamantium claws, which are his signature weapons for combat. In the image, I can see Wolverine with his claws extended from his knuckles. There are multiple claws visible on both his left and right hands, crossed in front of his chest.} (GRPO; prompt: ``claw'')}

\textcolor{grtogreen}{\textit{Looking at the image, I need to identify "Wolverine's body parts used for attacking enemies". This clearly refers to Wolverine's adamantium claws, which are his signature weapons and are extended from his knuckles. In the image, I can see Wolverine (played by Hugh Jackman) with his claws fully extended, held in a fighting stance. There are multiple claws visible on both his left and right hands.} (GRTO; prompt: ``claw'')}

\noindent\textbf{Figure~\ref{fig:qual_reasonseg}, row 4} (``In a bakery, bread and pastry products need to be cooled down after they come out of the oven. What area in the picture could be used for this purpose?'').

\textcolor{grpoblue}{\textit{Looking at the query, it's asking for an area in the picture that could be used for cooling bread and pastry products after they come out of the oven. In baking, a wire rack (also called a cooling rack) is specifically designed for this purpose - it allows air to circulate around the baked goods to cool them evenly without getting soggy. In the image, I can see several freshly baked pretzels resting on a metal grid structure. This structure has wire mesh with spaces between the wires, which is characteristic of a cooling rack. The pretzels are arranged on this rack, which is exactly where they would be cooled after baking.} (GRPO; prompt: ``cooling rack'')}

\textcolor{grtogreen}{\textit{Looking at the query, it's asking for an area in the picture that could be used to cool down bread and pastry products after they come out of the oven. In bakeries, this is typically done using a cooling rack, which allows air to circulate around the baked goods to cool them evenly without getting soggy. In the image, I can see several freshly baked pretzels arranged on a metal grid structure. This grid is clearly a cooling rack, designed specifically for this purpose. It has the characteristic wire mesh pattern that allows air to flow underneath and around the baked goods.} (GRTO; prompt: ``cooling rack'')}

\section{Error study}\label{sec:error_study}

\begin{figure*}[t]
    \centering
    \includegraphics[width=\linewidth]{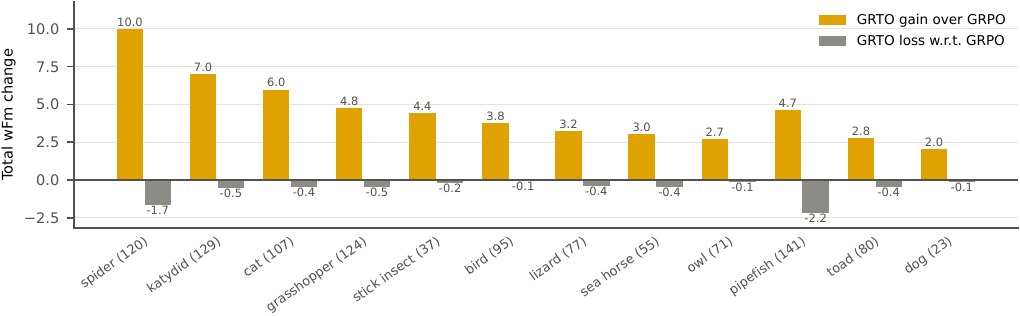}
    \caption{COD10K error study: total per-image weighted F-measure gain and loss of GRTO with respect to GRPO for the 12 classes with the largest net change (number of test images in parentheses).}
    \label{fig:err_cod10k}
\end{figure*}

\begin{figure*}[t]
    \centering
    \includegraphics[width=\linewidth]{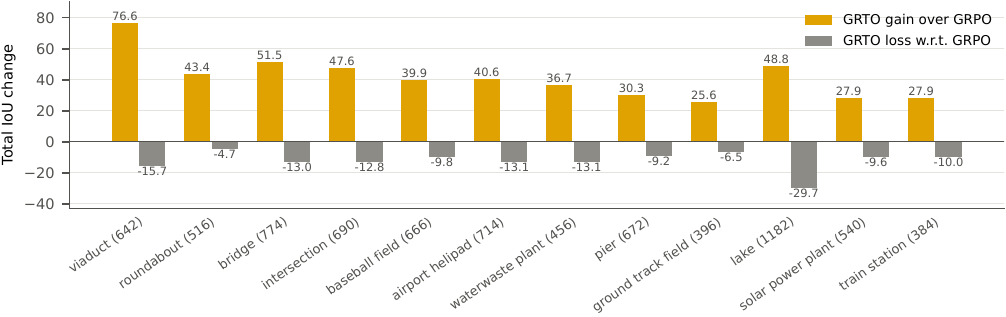}
    \caption{EarthReason error study: total per-question IoU gain and loss of GRTO with respect to GRPO for the 12 classes with the largest net change (number of test and validation questions in parentheses).}
    \label{fig:err_earthreason}
\end{figure*}

We analyze which classes benefit most from GRTO on camouflage detection and remote sensing; ReasonSeg provides no class annotations. For every test sample, we compute the difference between GRTO and GRPO in the per-sample metric (weighted F-measure for COD10K, IoU for EarthReason), averaged over the two GRTO seeds. For each class, we sum the positive differences (gain) and the negative differences (loss) separately. Figures~\ref{fig:err_cod10k} and~\ref{fig:err_earthreason} show the 12 classes with the largest net change. Classes are given by the COD10K file names and the EarthReason class annotations; EarthReason includes the test and validation splits.

GRTO improves most classes: the net change is positive for 59 of the 69 COD10K classes and 24 of the 28 EarthReason classes. On COD10K, the largest gains are on animals with thin or elongated structures that blend with their surroundings, such as spiders, katydids, grasshoppers, and stick insects, with only small losses. On EarthReason, they are on structured man-made targets such as viaducts, roundabouts, bridges, and intersections. Only few classes show a net loss: on COD10K, ten classes with a total of 96 of the 2,026 test images (4.7\%), the largest being leopard ($-$0.44) and crocodile ($-$0.27); on EarthReason, island, beach, stadium, and railway, with 2,598 of 18,016 questions (14\%) and net losses of $-$0.5 to $-$12.5, small compared with the largest net gains (\eg, $+$60.9 for viaduct).

\section{Prompts}\label{sec:prompts_used}

We give the exact prompts used to query the policy, reproduced verbatim. Each prompt is a single user turn preceded by the image, without a system prompt.

\noindent\textbf{Camouflage detection} (InternVL3.5-8B). The prompt is static: the same text is used for every image.

\begin{quote}
\small
\ttfamily
Please find the camouflaged human(s) or animal(s) with bbox(es). Also provide exactly one referential noun phrase that uniquely identifies the target. The noun phrase should specify species and any physical characteristics (color, texture, etc). The noun phrase should NOT include information about the target's surroundings - ONLY information about the target itself. When providing bboxes, double-check no targets are missed. One bbox should cover one instance of a target and contain all its visible parts.Output the final answer in <answer> </answer> tags. Output the bbox(es) of the interested object(s), along with the identifying noun phrase, in JSON format. i.e.:\\
<answer>\{"noun phrase": "brown mottled spider with bristly legs", "objects": [\{"bbox": [10,100,200,210]\}, \{"bbox": [225,296,706,786]\}]\}</answer>\\
This is an example only. Now it's your turn:
\end{quote}

\noindent\textbf{Remote sensing} (InternVL3.5-8B). \texttt{\{Question\}} is replaced by the EarthReason question.

\begin{quote}
\small
\ttfamily
Please find "\{Question\}" with bbox(es). Also provide exactly one referential noun phrase that uniquely identifies the objects within the image. Return ALL matching instances; if there are no matches, return an empty answer (<answer></answer>). double-check none are missed. Output the bbox(es) of the interested object(s), along with a noun phrase, in JSON format. i.e., <answer>\{"noun phrase": "a player in red", "objects": [[10,100,200,210], [225,296,706,786]]\}</answer>
\end{quote}

\noindent\textbf{Reasoning segmentation} (StAR, Qwen3-VL-8B). We use the StAR prompt~\cite{yun2026star}, where \texttt{\{Question\}} is the lower-cased query without trailing punctuation.

\begin{quote}
\small
\ttfamily
Please find "\{Question\}" with bbox(es) and point(s). Also provide a short label for each object. First, understand and summarize what the query ---"\{Question\}"--- is likely referring to (which object or concept). Then apply this to the image and find the matched target object(s). Return ALL matching instances; if there are no matches, return an empty list (<answer>[]</answer>). double-check none are missed. Output the thinking process in <think> </think> and final answer in <answer> </answer> tags. Output the bbox(es) and point(s) inside the interested object(s), along with a short label, in JSON format. i.e., <think> thinking process (step-by-step reasoning) here </think> <answer>[\{"label": "chair", "bbox\_2d": [10,100,200,210], "point\_2d": [30,110]\}, \{"label": "train track", "bbox\_2d": [225,296,706,786], "point\_2d": [302,410]\}]</answer>
\end{quote}

\end{document}